\documentclass[journal]{IEEEtran}

\usepackage{mathrsfs}  %%%%%%%%%%%%%%%%%%%
\usepackage{graphicx}
\usepackage{subfloat}
\usepackage{float}
\usepackage{amsmath}
\usepackage{CJK}
\usepackage{amssymb}
\usepackage{indentfirst}
\usepackage{mathtools}
\usepackage{subfigure}
\usepackage{stfloats}
\usepackage{mathrsfs}
\usepackage{bm}
\usepackage{blindtext}
\usepackage{algorithm}
\usepackage{algpseudocode}
\usepackage{amsmath}
\usepackage{amsmath,amssymb}
\DeclareSymbolFont{EulerExtension}{U}{euex}{m}{n}
\DeclareMathSymbol{\euintop}{\mathop} {EulerExtension}{"52}
\DeclareMathSymbol{\euointop}{\mathop} {EulerExtension}{"48}

\usepackage[Symbol]{upgreek}
\usepackage{booktabs}
\usepackage{multirow}

\usepackage{array}
\newcommand{\PreserveBackslash}[1]{\let\temp=\\#1\let\\=\temp}

\ifCLASSINFOpdf

\else

\fi

\begin{document}
\title{Spatial and Temporal Consistency-aware Dynamic Adaptive Streaming for 360-degree Videos}

\author{Hui~Yuan,~\IEEEmembership{Senior Member,~IEEE,}
        Shiyun~Zhao,
        Junhui~Hou,~\IEEEmembership{Member,~IEEE,}\\
        Xuekai~Wei,
        and~Sam~Kwong,~\IEEEmembership{Fellow,~IEEE}% <-this % stops a space
%\thanks{Manuscript Received.}% <-this % stops a space
\thanks{This work was supported in part by the National Natural Science Foundation of China under Grants 61571274 and 61871342; in part by the Shandong Natural Science Funds for Distinguished Young Scholar under Grant JQ201614; in part by the National Key R\&D Program of China under Grants 2018YFC0831003; in part by the Young Scholars Program of Shandong University (YSPSDU) under Grant 2015WLJH39. \emph{The corresponding author is Junhui Hou.}}% <-this % stops a space
\thanks{H. Yuan is with the School of Control Science and Engineering, Shandong University, Ji'nan 250061, China (Email: huiyuan@sdu.edu.cn).}% <-this % stops a space
\thanks{S. Zhao is with the School of Information Science and Engineering, Shandong University, Qingdao, 266237, China (Email: zhaoshiyun@yeah.net).}% <-this % stops a space
\thanks{J. Hou, X. Wei and S. Kwong are with the Department of Computer Science, City University of Hong Kong, Kowloon, Hong Kong (Email: jh.hou@cityu.edu.hk, xuekaiwei2-c@my.cityu.edu.hk, and cssamk@cityu.edu.hk).}
\thanks{\textbf{Copyright (c) 2019 IEEE. Personal use of this material is permitted. However, permission to use this material for any other purposes must be obtained from the IEEE by sending an email to pubspermissions@ieee.org.}}
}
%%%%%%%%%%%%%%%%%%%%%%%%%%%%%%%%%%%%%% The paper headers%%%%%%%%%%%%%%%%%%%%%%%%%%%%%%%%%%%%%%%%

\maketitle

\begin{abstract}
The 360-degree video allows users to enjoy the whole scene by interactively switching viewports. However, the huge data volume of the 360-degree video limits its remote applications via network. To provide high quality of experience (\emph{QoE}) for remote web users, this paper presents a tile-based adaptive streaming method for 360-degree videos. First, we propose a simple yet effective rate adaptation algorithm to determine the requested bitrate for downloading the current video segment by considering the balance between the buffer length and video quality. Then, we propose to use a Gaussian model to predict the field of view at the beginning of each requested video segment. To deal with the circumstance that the view angle is switched during the display of a video segment, we propose to download all the tiles in the 360-degree video with different priorities based on a Zipf model. Finally, in order to allocate bitrates for all the tiles, a two-stage optimization algorithm is proposed to preserve the quality of tiles in FoV and guarantee the spatial and temporal smoothness. Experimental results demonstrate the effectiveness and advantage of the proposed method compared with the state-of-the-art methods. That is, our method preserves both the quality and the smoothness of tiles in FoV, thus providing the best \emph{QoE} for users.
\end{abstract}


\IEEEpeerreviewmaketitle

%%%%%%%%%%%%%%%%%%%%%%%%%%%%%% Introduction %%%%%%%%%%%%%%%%%%%%%%%%%%%%%%%%%%%%%

\section{Introduction}
\IEEEPARstart{W}{ith} great advances in multimedia and computer technologies, augmented and virtual reality (AR/VR) are becoming more and more popular in both academic and industrial communities [1]. As one kind of immersive media for representing AR/VR scenes, the 360-degree video can provide users with more immersive experience than traditional monoscopic videos. By making use of various head-mounted displays (HMDs) that can detect head movements of users and provide corresponding viewport for users, such as Samsung Gear VR Glass, Oculus Rift, and HTC Vive, users can enjoy the scene as if they were there. However, due to the huge data volume and the complex rendering algorithms, it is difficult to achieve all the functions (e.g. storage, rendering, interaction, etc.) of a VR system for a mobile device. Therefore, the scenario that a user interacts with a remote webserver (such as the edge computing node and content distribution server) to enjoy VR/AR applications is taken into account. In this case, users will interactively request video content of different viewports from the webserver. The function of the webserver is to storage 360-degree videos and send video content to users based on their requirements. Due to the diverse network environments between users and the webserver, how to efficiently transmit 360-degree videos to users is becoming a critical problem.

To deal with diverse network environments, adaptive streaming techniques, e.g. Hypertext Transfer Protocol (HTTP) streaming, are becoming more and more popular in these years [2][3]. Under this circumstance, the Moving Picture Experts Group (MPEG) of International Standardization Organization (ISO)/International Electro technical Commission (IEC) has standardized a protocol named Dynamic Adaptive Streaming over HTTP (DASH) [4][5]. Owing to its highly adaptive property, DASH is a reliable solution for real-time 360-degree video transmission.

In a DASH-based video delivery system, media content is first divided into several segments (or chunks) [6] with the same playback duration, e.g., 2 seconds. Then, each segment is encoded with different bitrates corresponding to different quality levels and stored in a webserver. The webserver will generate a manifest (MPD) file that records the description of all the available segments of a video, e.g., URL addresses, segment lengths, quality levels, resolutions, etc. The user will request to download segments with different bitrates from the webserver to adapt to the network throughput variation according to the received MPD file, user preference, etc.

Considering the fact that users can only see the field of view (FoV) that contains the current region of interest (ROI) of users [7], the server allows to transmit a part of a 360-degree video to the user. Fortunately, there is a ``tile'' concept that can support to divide a high resolution video into several parts in the H.265/High Efficiency Video Coding (HEVC) video coding standard [8]. Therefore, in addition to dividing video content into segments along the temporal axis, we can also divide the 360-degree video into multiple tiles spatially to satisfy user's viewing preference adaptively. Each tile is then encoded independently into different versions with multiple bitrates and stored in the webserver. The webserver will deliver the tiles that contain the current FoV with high quality according to the user's request so as to provide higher quality of experience (\emph{QoE}) with low bandwidth consumption for users. Tile-based 360-degree video streaming can provide a variety of viewpoints for users adaptively, but it may also suffer from high viewport-switching delay. The reason is that there is no video content of the other viewports in the currently delivered video content. During the display time of the current video segment, delay occurs when the user changes his/her viewport suddenly. Therefore, not only the tiles that contain the current FoV but also the other tiles should be transmitted to the user. To save bandwidth consumption, the low bitrate versions of the tiles that may not be viewed can be delivered. Besides, because of the time and space discrete characteristics of tile-based 360-degree video, the effects of spatial and temporal smoothness on \emph{QoE} should also be considered [9][10] during the delivery. Consequently, the challenges of adaptively streaming a tile-based 360-degree video lie in

(a) how to design a simple yet effective rate adaptation algorithm to tackle the network throughput variation;

(b) how to predict the view angle at the beginning of each request, and the priority of each tile in the current requested video segment and

(c) how to find the best bitrate combination of all the tiles by considering not only the video quality but also the spatial and temporal smoothness to cope with sudden view switching in the display duration of a video segment.

Accordingly, to improve user \emph{QoE}, we first propose a rate adaptation algorithm to adapt to the diverse network throughput. Then, a Gaussian model [11] and a Zipf model [12] are used to predict view angle at the beginning of each request and the priority of each tile during the display time of a video segment. Finally, we model the bitrates combination problem as an optimization problem by considering both the video quality and the spatial-temporal smoothness of FoVs.

The rest of this paper is organized as follows. In Section II, basic concepts and related work are briefly reviewed. Then, the proposed rate adaption algorithm, view angle prediction, and the solution of the bitrates combination problem are given in Section III. Experimental results and analyses are provided in Section IV. Finally, Section V concludes this paper.

%%%%%%%%%%%%%%%%%%%%%%%%%%%%%% II %%%%%%%%%%%%%%%%%%%%%%%%%%%%%%%%%%%%%
\section{Basic Concepts and Related Work}
In this section the basic concepts of the tile-based 360-degree video and the DASH system are reviewed first. Then, the related work of rate adaptation and bit allocation algorithms are briefly introduced.
\subsection{Tile-based 360-degree Video}  % II――A
The 360-degree video is usually captured by a set of cameras that are fixed on a sphere or a circle uniformly [13]. After aggregating and stitching [14] the images captured by different cameras, a panoramic image can be generated, and finally formed to be a 360-degree video. Because it is hard to store and operate the 360-degree video in the spherical domain, the 360-degree video is usually projected onto a two-dimensional (2D) plane [15][16], and represented by the commonly used equirectangular format (ERP) [17][18]. With the help of HMDs, users will enjoy the spherical video content by re-projecting the ERP formatted images onto a sphere. The resolution of an ERP image is usually at least 3840$\times$1920 (4K). Limited by the FoV (usually, 90-degree and 110-degree in the vertical and horizontal directions, respectively), users can only enjoy a part of the whole 360-degree video at a time [19]. Therefore, to deal with the huge data volume, it is better to process and transmit a part of the 360-degree video.

In H.265/HEVC [8][20], the concept of tile is to divide a video picture into regular-sized, rectangular regions which can be independently encoded and decoded in order to enable parallel processing architectures and spatial random access to local regions. The tile-based adaptive streaming is a suitable way for transmitting 360-degree videos as the visible region of the video is only a small part of the whole video content. As shown in Fig. 1, after projecting the sphere onto a 2D plane, a 360-degree video picture with ERP format is divided into 24 tiles with 4 rows and 6 columns, the corresponding area of FoV is shown by the red block. When the user changes his/her head to another direction from the current view angle, the FoV will be switched accordingly. Different tile partitions result in different panoramic experience. Existing studies have presented efficient methods of how to cut a 360-degree video into suitable sized tiles [21]-[24]. By investigating the tradeoff between bitrate overhead, view switching adaptivity, and the bandwidth consumption, the partition method with 4 rows and 6 columns is the recommended [20].
\begin{figure}
\centering
\includegraphics[width=8.5cm]{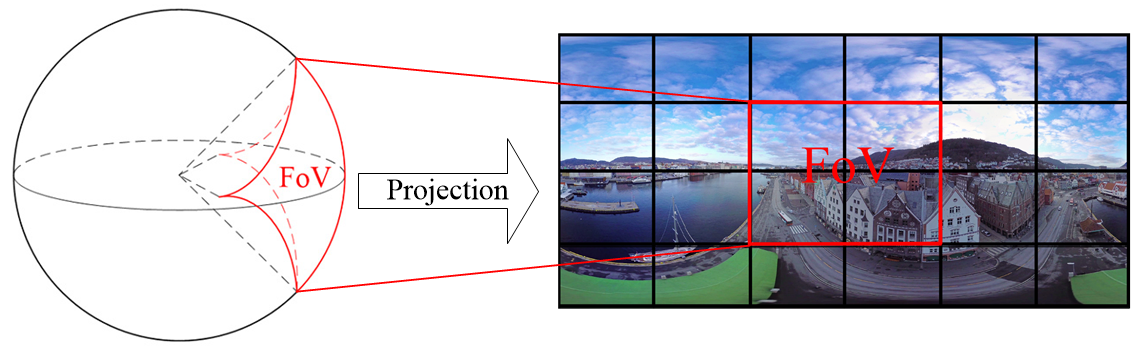}
\caption{An example of the tile-based 360-degree video.}
\label{fig1}
\end{figure}
\subsection{DASH System}  % II――B
In a DASH system, videos are encoded into different bitrates and divided into several segments with the same display duration, e.g. 2 seconds, for the purpose of satisfying user requirement. For the tile-based 360-degree video, as shown in Fig. 2, a 360-degree video is temporally divided into segments and then spatially cut into tiles. The tiles are then encoded with different bitrates by an encoder, e.g., H.265/HEVC. The webserver in the DASH system will store the encoded stream of each tile, and generate a corresponding MPD file that records URL address of each stream, segment length (represented by display time), spatial resolutions, quality levels (corresponding to different bitrate versions), and the spatial relationship description (SRD) [23] which describes the spatial information of rectangular tiles in a video content so as to locate the FoV accurately.

\begin{figure}
\centering
\includegraphics[width=9cm]{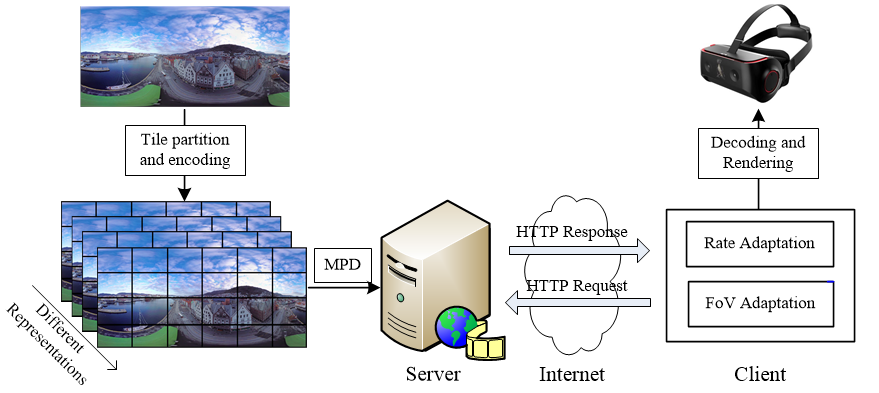}
\caption{DASH system architecture of a tile-based 360-degree video streaming.}
\label{fig2}
\end{figure}
The user client first downloads the MPD file by using HTTP protocol from the server. Then it will parse the MPD file and request the tiles with desired quality levels according to the buffer length, available throughput and user's FoV. In this procedure, the delivery service has to adapt to the view switching of the user as well as the network throughput fluctuation. Finally, the received tiles are decoded and used to render the visual content with the help of HMDs.
\subsection{Rate Adaptation Algorithms}  % II――C
To achieve network throughput adaptation and decrease the number of display interruptions, rate adaption algorithms are usually designed based on the buffer occupancy and throughput estimation [25]. In brief, there are two main kinds of bitrate adaptation algorithms, i.e., quality-first-based algorithm (QFA) and buffer-first-based algorithm (BFA).

The QFA [26][27] tries to download video segments with relatively high available video quality representation (i.e., bitrate) to match the predicted bandwidth no matter how large the buffer is. However, this kind of algorithm will lead to frequent quality fluctuations, which is extremely annoying to users [28].

In contrast, the BFA [29][30] aims to remain the buffer length stable so as to ensure continuous and smooth video playback. However, the received video quality levels in this algorithm are not taken in first consideration. The user client always plans to fill a predefined buffer length with the lowest video quality level. When the buffer length is larger than a predefined buffer length, it tries to request video segments with the next higher quality representation until buffer length is lower than the predefined buffer length. The BFA prefers to fill its predefined buffer length before switching to higher bitrates. Thus, it avoids the risk of inaccurate throughput estimation, and stabilizes the buffer so as to ensure smooth video playback. But, BFA usually downloads video segments with significantly lower qualities than the QFA, causing abundant bandwidth waste.

Hence, there should be a trade-off between QFA and BFA. A detailed review of the rate adaptation algorithms for adaptive streaming is shown in [31]. On the basis of QFA and BFA, we also propose a simple but effective rate adaptation algorithm in \textbf{\emph{Part A}} of \textbf{Section III}.
\subsection{Bit Allocation Algorithms}  % II――D
Besides rate adaptation algorithm, bit allocation algorithm is another critical technology that can affect the user \emph{QoE}. For a tile-based 360-degree video, the problem can be described as how to optimally allocate a target bitrate to all the tiles so as to guarantee the quality of the FoV as well as spatial and temporal smoothness of different tiles.

\emph{Yun and Chung} [32] proposed a bit allocation algorithm to minimize the view-switching delay by employing a buffer control, parallel streaming, and server push scheme. In this algorithms, all the possible views that may be switched to are equally treated. \emph{Ban et al.} [33] proposed a \emph{QoE}-driven optimization framework under limited network for tile-based adaptive 360-degeree video streaming. The bitrates of all the tiles are determined optimally, aiming at maximizing the overall quality and minimizing the spatial and temporal quality variation of all the tiles without considering the FoV.

Because users can only see the video content in FoV, most researchers focused on how to allocate bitrates for tiles by considering the priorities of tiles in FoV. \emph{Rossi and Toni} [34] proposed an algorithm to choose the best set of tiles by solving an integer linear programming problem. \emph{Chao et al.} [35] proposed an optimal bitrate allocation scheme for a novel multiple views navigation rule, allowing the clients to maximize the video quality over DASH. \emph{Xie et al.} [36] modeled the perceptual impact of the quality variations (through adapting the quantization step-size and spatial resolution) with respect to the refinement duration, and yield a product of two closed-form exponential functions. \emph{Yu et al.} [37] proposed a convolutional neural network (ConvNet) assisted seamless multi-view video streaming system. In this framework, ConvNet assisted multi-view representation algorithm and a bit allocation mechanism guided by a navigation model were combined to provide flexible interactivity and seamless navigation under network bandwidth fluctuations without compromising on multi-view video compression efficiency. \emph{Carlsson et al.} [38] proposed a novel multi-video stream bundle framework for interactive video playback that allows users to dynamically switch among multiple parallel video streams capturing the same images from different viewpoints. \emph{Liu et al.} [19] described a server-side rate adaptation strategy for 360-degree video streaming, in which a tile visibility probability model is established. \emph{Yang et al.} [39] proposed a DASH-based 360-degree video adaptive transmission algorithm based on user's viewport. In this algorithms, the network throughput, video buffer length, and viewport were considered for adaptive decision so that it could effectively save the bandwidth. \emph{Zhang et al.} [40] designed a streaming transmission strategy to select the optimal set of views for users to download, such that the navigation quality experienced of the user can be optimized with the bandwidth constraint.

To sum up, these algorithms try to improve user experience from aspects of average quality, display continuity, quality variation, etc. Due to the space and time partition of 360-degree videos in tile-based adaptive streaming, how to preserve the spatial and temporal smoothness for FoVs is still a challenge.
%%%%%%%%%%%%%%%%%%%%%%%%%%%%%% III %%%%%%%%%%%%%%%%%%%%%%%%%%%%%%%%%%%%%
\section{The Proposed Method}
As shown in Fig. 3, in a tile-based 360-degree video adaptive streaming system, a 360-degree video with the ERP format is first temporally partitioned into $L$ segments with the same display duration (e.g., 2 seconds). Each segment is then cut into $N$ tiles spatially. Consequently, each tile is encoded into $U$ quality levels (or bitrate versions), and the corresponding bitrate of the $n$-th tile with the $u$-th quality level in the $l$ -th segment is denoted as ${R_{l,n,u}}$, $l \in\{1, \cdots, L\}$, $n \in \left\{ {1, \cdots } \right.,N\} ,u \in \left\{ {1, \cdots } \right.,U\}$. Owing to the additional spatial division, the representation of a tile-based 360-degree video is more flexible than that of a traditional video that is only temporally divided. In this system, the user clients will first determine the requested bitrate based on the buffer length and the predicted throughput, and then download the video segment whose bitrate is the closest to the requested bitrate. Since a video segment is divided into several tiles, the combination of those tiles with different bitrates must be determined so as to provide the highest \emph{QoE} under the constraint of the requested bitrate.

\begin{figure}
\centering
\includegraphics[width=8.5cm]{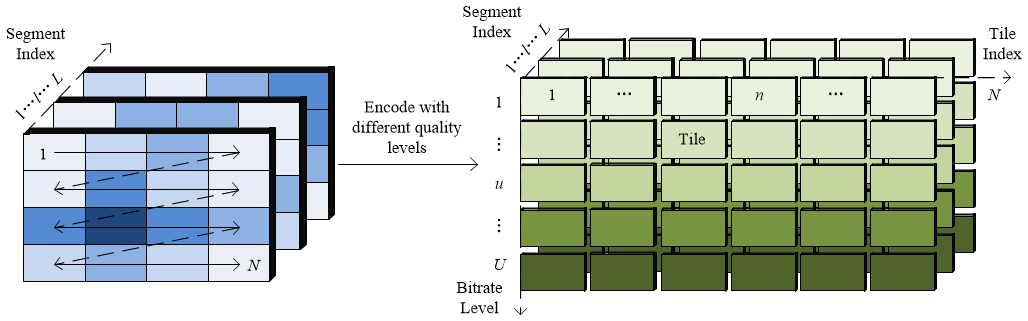}
\caption{The storage structure of tile-based videos for adaptive streaming.}
\label{fig3}
\end{figure}
In the following, as shown in Fig. 4, we will first present a simple yet effective rate adaption algorithm. Then, we propose to use a Gaussian distribution-based  FoV switching model to predict the view angle trajectory of each segment and Zipf model-based tile priority model to determine the priority of each tile in a segment. After that, we propose a bit allocation algorithm to find the best combination of tiles based on the tile priorities.
\begin{figure}
\centering
\includegraphics[width=5cm]{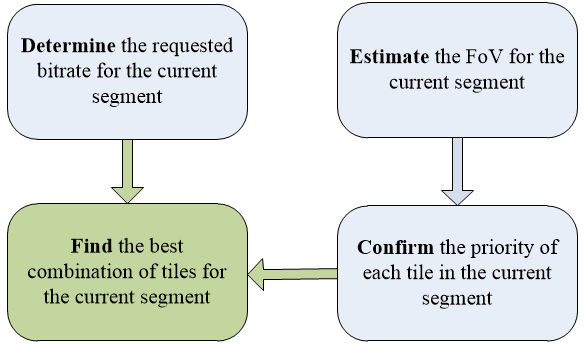}
\caption{System diagram of the proposed method.}
\label{fig4}
\end{figure}
\subsection{Buffer-quality-based Rate Adaptation Algorithm}     % III――A
Inspired by QFA [26][27] and BFA [29][30], we propose a simple yet effective rate adaption algorithm, namely buffer-quality-based algorithm (BQA), which considers both the buffer length and video quality to guarantee the quality and smoothness of the received videos. Since the rate adaptation algorithm is independent of the video content, we explain it by taking a video without the tile partition for simplicity.

%Let $l \in \left\{ {1, \cdots } \right.,L\}$ denote the index of video segments, $\nu \in \left\{ {1, \cdots ,{\cal V}} \right\}$ the quality level index of a segment, and

Let ${b_{cur}}$ denote the current buffer length. In the proposed algorithm, we set two buffer thresholds (${b_{min}}$ and ${b_{max}}$). At the beginning, the user client will request the video segments with the lowest quality to quickly fill the buffer until the buffer length is equal to ${b_0}$ which is usually set to be 2 seconds. Then, at a certain decision time, the user client will first predict the throughput ${T_{cur}}$ by the download time and total bitrate of previously downloaded ${L_0}$ video segments:
\begin{equation}\label{E1}
   {T_{cur}} = \frac{1}{{{L_0}}}\sum\nolimits_{l - {L_0} + 1}^l {\frac{{{r_l} \cdot {t_0}}}{{{t_{download,l}}}}},
\end{equation}
where ${r_l}$ and ${t_{download,l}}$ denote the bitrate and the download time of the $l$-th segment, respectively. And the playback duration of a segment is equal to ${t_0}$ (e.g., 2 seconds). The user client will then calculate the segment quality level (${\emph{u}_{cur}}$) whose corresponding bitrate is less than and closest to ${T_{cur}}$. When ${b_{cur}} < {b_{min}}$, the user client will request the video segment with the next lower quality level than ${\emph{u}_{cur}}$; when ${b_{cur}} > {b_{max}}$, the user client will request the video segment with the next higher quality level than ${\emph{u}_{cur}}$; otherwise, when ${b_{min}} \le {b_{cur}} \le {b_{max}}$, the user client will request the video segment with the quality level ${\emph{u}_{cur}}$.  According to the buffer length and predicted throughput capacity, we can get the final selected segment quality level ${\emph{u}_{select}}$ and the corresponding bitrate ${{\rm{\mathbb{R}}}_{l,request}}$. In summary, the proposed rate adaption algorithm first guarantees the minimal buffer length, and then controls the buffer length such that it fluctuates between the two predefined buffer thresholds ${b_{min}}$ and ${b_{max}}$. \textbf{\emph{Algorithm 1}} shows the detailed procedure of our proposed rate adaptation method.
% '算法伪代码1'
\begin{table}[htbp]
  \centering
    \begin{tabular}{l}
    \toprule
    \textbf{\textit{Algorithm 1}: Rate Adaptation Algorithm for Video Segment Request} \\
    \midrule
        \ 1: Initially, the user client requests the minimum bitrate from the \\
          \quad \ server to quickly establish the initial buffer length $b_{0}$.\\
        \ 2: \textbf{while} $l\leq L$ \textbf{do}\\
        \ 3:\ \quad predict throughput $T_{cur}$ and calculate quality level $\emph{u}_{cur}$\\
        \ 4:\ \quad \textbf{if} $b_{cur}<b_{min}$ \textbf{then}\\
        \ 5:\ \qquad $\emph{u}_{select} = \emph{u}_{cur}-1$\\
        \ 6:\ \quad \textbf{else if} $b_{min}\leq b_{cur} \leq b_{max}$ \textbf{then}\\
        \ 7:\ \qquad $\emph{u}_{select} = \emph{u}_{cur}$\\
        \ 8:\ \quad \textbf{else}\\
        \ 9:\ \qquad $\emph{u}_{select} = \emph{u}_{cur}+1$\\
       10:\ \quad \textbf{end if}\\
       11:\ \quad $l=l+1$\\
       12: \textbf{end while}\\
    \bottomrule
    \end{tabular}
  \label{tab:addlabel}
\end{table}

For the tile-based 360-degree video, as there are no explicit quality levels of a segment, we have to define a rule to confirm the requested bitrate. Based on \textbf{\emph{Algorithm 1}}, when ${b_{cur}} < {b_{min}}$, the requested bitrate ${{\rm{\mathbb{R}}}_{l,request}}$ should be lower than throughput ${T_{cur}}$; when $b_{min}\leq b_{cur} \leq b_{max}$, the requested bitrate ${{\rm{\mathbb{R}}}_{l,request}}$ should match to ${T_{cur}}$; when ${b_{cur}} > {b_{max}}$, ${{\rm{\mathbb{R}}}_{l,request}}$ should be higher than ${T_{cur}}$. We define the following function to calculate ${{\rm{\mathbb{R}}}_{l,request}}$:
\begin{equation}\label{E2}
   {\mathbb{R}_{l,request}} = \varepsilon  \cdot {T_{cur}},
\end{equation}
where $\varepsilon$ is the coefficient depending on the current buffer length:
\begin{equation}\label{E3}
   \varepsilon  = \left\{ {\begin{array}{*{20}{c}}
{\frac{{{b_{cur}}}}{{{b_{min}}}}\qquad \qquad {b_{cur}} < {b_{min}}},\vspace{1ex}\\
{1\qquad {b_{min}} \le {b_{cur}} \le {b_{max}}},\vspace{1ex}\\
{\frac{{{b_{cur}}}}{{{b_{max}}}}\qquad \qquad {b_{cur}} > {b_{max}}}.
\end{array}} \right.
\end{equation}

\subsection{FoV Switching and Tile Priority Model}     % III――B
For remote applications, to deal with the limited and varied network environment, the user client must pre-fetch enough bit streams to fill the buffer. At the beginning (buffer is empty), the FoV can be calculated directly by the HMD of the user. After that, when the buffer is not empty, it is very important to predict the FoV accurately [41]-[43]. Existing FoV estimation methods can be roughly classified into three categories, i.e., data driven approaches [44], probability model based approaches [35][45] and motion saliency detection based approaches [46].  Although data driven approaches and motion saliency detection based approaches achieve good performance, the viewport movement depends only on the subjective will of a user, and it can never be predicted accurately. Because immersion also depends on audio and motion perception apart from video content [47]-[49]. In addition, the implementation complexity of these approaches is also high.

In this paper, from the perspective of a practical application, we choose the probability model based approach to predict the FoV of users, owing to its low computational complexity. Specifically, the Gaussian model [11] is employed to determine the FoVs at the beginning of each video segment. First, we define 20 FoV patterns as shown in Fig. 5 by considering the projection relationship between the spherical structure and the ERP format. When requesting video segments, the variation of the FoV pattern follows the Normal distribution, i.e., ${\cal N}(\mu ,{\sigma ^2})$, where the mean $\mu$ and variance ${\sigma ^2}$  reflect the mean value and dispersion degree of the FoV pattern variation, respectively. Based on the assumption that the center of a picture is usually the ROI, we restrict the mean value of the Gaussian model to be 11 corresponding to the 11-th FoV pattern, as shown in Fig.5. It is worth pointing out the mean value can also be defined based on other ROI or saliency detection algorithms. The Gaussian model will generate a random value ranging from 1 to 20 to determine the FoV of a video segment, as shown in Fig. 6.
%${\sigma ^2}$=4和${\sigma ^2}$=9.

\begin{figure}
\centering
\includegraphics[width=9cm]{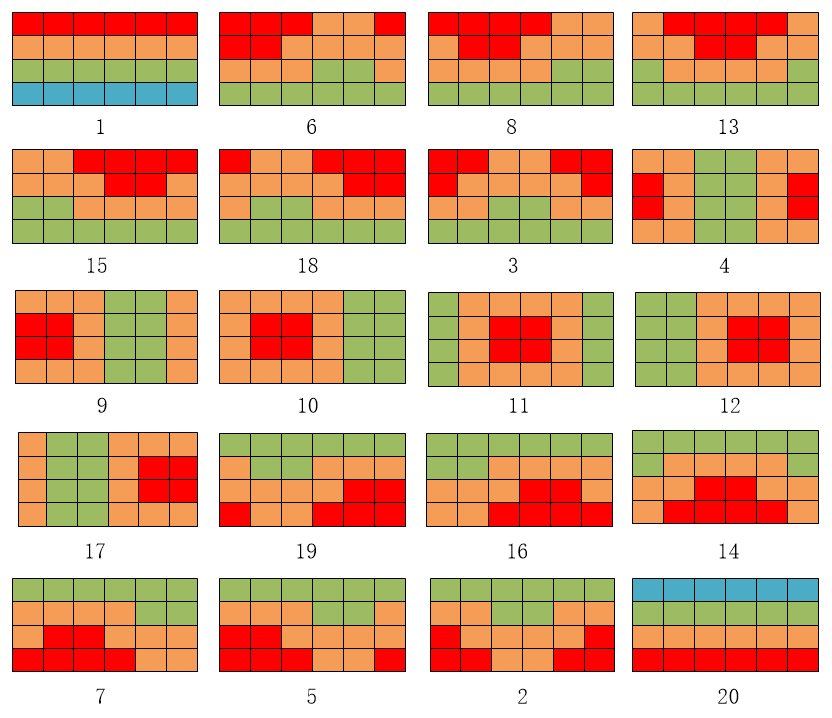}
\caption{FoV patterns,the red area denotes the FoV of each pattern.}
\label{fig5}
\end{figure}

%\begin{figure}
%\centering
%\includegraphics[width=6cm]{fig6.png}
%\caption{A Gaussian Model-based FoV trajectory example.}
%\label{fig6}
%\end{figure}

\begin{figure}
\centering
\subfigure[]{
\label{fig6:subfig:a} %% label for first subfigure
\includegraphics[width=4.25cm]{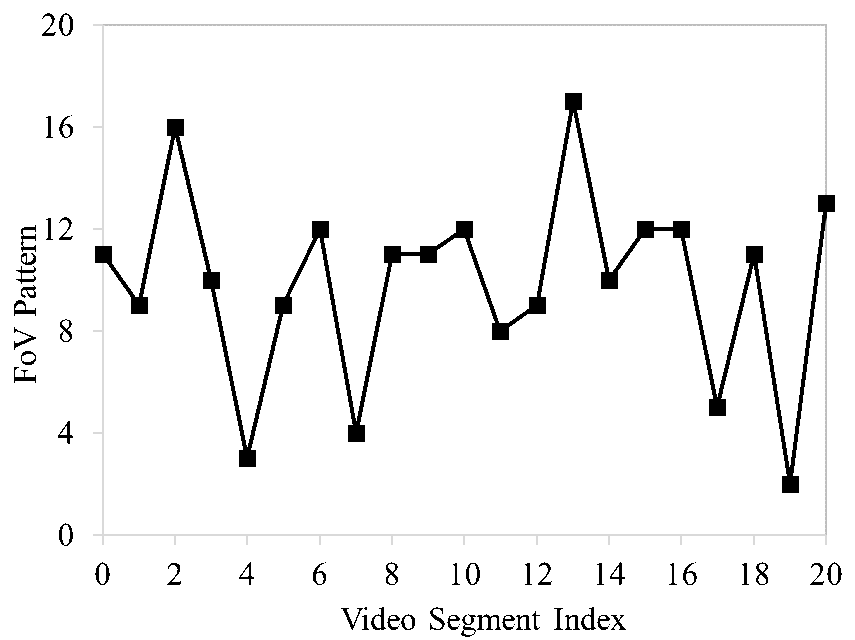}}
\subfigure[]{
\label{fig6:subfig:b} %% label for second subfigure
\includegraphics[width=4.25cm]{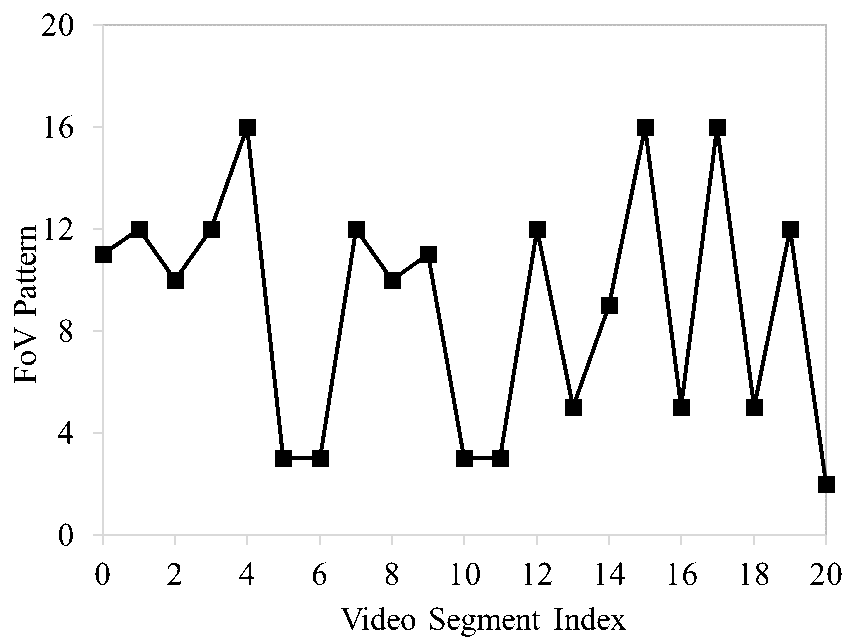}}
\caption{Gaussian model-based FoV trajectory examples, (a) ${\sigma ^2}$=4, (b) ${\sigma ^2}$=9.}
\label{fig6}
\end{figure}

To deal with the unpredictable change of the view angle during the display time of a video segment, the webserver has to transmit all tiles to the user. In order to save bandwidth, a Zipf model [12] is used to calculate the priority of each tile in a video segment. Thus, the high (resp. low) quality versions of the tiles with high (resp. low) priorities can be transmitted under the constraint of the requested bitrate.

First, we set 4 priorities for tiles according to the distance from FoV to each tile. As shown in Fig. 5, different colors represent tiles with different priorities, which are \emph{red}, \emph{orange}, \emph{green} and \emph{blue}, respectively. We assume the priority of a tile decreases with its distance to the FOV. Therefore, the priorities of \emph{red}, \emph{orange}, \emph{green} and \emph{blue} regions are in a descending order. In order to quantitatively calculate the priority of each region, the Zipf model is used. Let  ${p_{l,c,\varphi }}$ denote the priority of the $\varphi$-th tile in a color region $c$, where $c \in \textbf{\emph{C}}=\left\{ {``red",``orange",``green" \emph{and}\ ``blue"} \right\}$ in $l$-th segment, and ${\pi _c}$ be the number of tiles in the color region $c$. Note that ${\pi _c}$ depends on the FoV patterns shown in Fig. 5, and the priority of each tile in the same color region is the same. Accordingly, from the basis of Zipf model, the priority of each tile can be calculated by solving
\begin{equation}\label{E4}
   \sum\nolimits_{c \in \textbf{\emph{C}}} {{\pi _c} \cdot } {p_{l,c,\varphi }} = 1,
\end{equation}
where
\begin{equation}\label{E5}
   {p_{l,c,\varphi }} = d_{l,c}^{ - 1},
\end{equation}
and ${d_{l,c}}$ denotes the relative distance from the color region $c$ to the \emph{red} region (FoV). To solve Eq. (4), we empirically set ${d_{l,``orange"}} = 2 \cdot {d_{l,``red"}}$, ${d_{l,``green"}} = 3 \cdot {d_{l,``red"}}$, and ${d_{l,``blue"}} = 4 \cdot {d_{l,``red"}}$.

\subsection{Bit Allocation Algorithm for Tiles}      % III――C
Based on the requested bitrate ${\mathbb{R}_{l,request}}$, the predicted FoV and the priority (${p_{l,n}},{\rm{ }}n \in \left\{ {1, \cdots ,N} \right\}$) of the $n$-th tile in the $l$-th segment, the remaining problem is how to determine the bitrate combination of tiles to improve the user \emph{QoE}, which can be formulated as a bit allocation problem, i.e., how to allocate appropriate bitrates for all the tiles with the constraint of a total target bitrate (the requested bitrate). Here, we propose a coarse-to-fine algorithm to solve the problem.

    \emph{Stage 1: Coarse Bit Allocation}

	In the first stage, we use the weighted distortion of tiles in the $l$-th requested segment as the objective function, and model the problem as
\begin{equation}\label{E6}
   \mathop {\min }\limits_{{\mathbb{R}_{l,n}}} \sum\nolimits_{n = 1}^N {\left( {{p_{l,n}}{\mathbb{D}_{l,n}}} \right)} ,\quad s.t.{\rm{ }}\sum\nolimits_{n = 1}^N {{\mathbb{R}_{l,n}} \le {\mathbb{R}_{l,request}},}
\end{equation}
where ${\mathbb{R}_{l,n}}$ and ${\mathbb{D}_{l,n}}$ denote the allocated bitrate and the corresponding distortion of the $n$-th tile, respectively.

To solve the problem, the rate distortion function must be explicitly determined. To find the analytic solution, mean squared error (MSE) is used as the distortion metric. According to the Cauchy distribution-based rate distortion function [50][51], the relationship between distortion ${\mathbb{D}_{l,n}}$ and rate ${\mathbb{R}_{l,n}}$ can be written as
\begin{equation}\label{E7}
   {\mathbb{D}_{l,n}} = {\alpha _{l,n}} \cdot {\mathbb{R}_{l,n}}^{ - {\beta _{l,n}}},
\end{equation}
where ${\alpha _{l,n}}$, ${\beta _{l,n}}>0$ are model parameters that can be obtained at the step of video encoding.

Since the Cauchy-based rate distortion function is convex, the constrained optimization problem of (6) can be converted into an unconstrained optimization problem by Lagrange multiplier method:
\begin{equation}\label{E8}
   \mathop {\min J}\limits_{{\mathbb{R}_{l,n}}} = \sum\nolimits_{n = 1}^N {\left( {{p_{l,n}}{\mathbb{D}_{l,n}}} \right) + \lambda \left( {\sum\nolimits_{n = 1}^N {{\mathbb{R}_{l,n}} - {\mathbb{R}_{l,request}}} } \right)} ,
\end{equation}
and solved by Karush-Kuhn-Tucher (KKT) conditions:
\begin{equation}\label{E9}
   \left\{ {\begin{array}{*{20}{c}}
{\begin{array}{*{20}{c}}
{\frac{{\partial J}}{{\partial {\mathbb{R}_{l,1}}}} = 0}\vspace{1ex}\\
{\frac{{\partial J}}{{\partial {\mathbb{R}_{l,2}}}} = 0}
\end{array}}\vspace{1ex}\\
 \cdots \vspace{1ex}\\
{\frac{{\partial J}}{{\partial {\mathbb{R}_{l,N}}}} = 0}\vspace{1ex}\\
{\frac{{\partial J}}{{\partial \lambda }} = 0}
\end{array}} \right.,
\end{equation}
where $\lambda$ is the Lagrange multiplier.

    \emph{Stage 2: Fine Bit Allocation}

It is worth pointing out that the resulting ${\mathbb{R}_{l,n}}$ from stage 1 may not be equal to any stored bitrate versions of tiles, i.e., ${R_{l,n,u}},u \in \left\{ {1, \cdots ,U} \right\}$. Therefore, the tiles whose bitrate versions are closest to, but not larger than ${\mathbb{R}_{l,n}}$ are selected first. For the selected tiles with certain quality levels, the corresponding bitrates are denoted by $\mathbb{R}_{l,n}^0$, which is the solution of Eq. (6). In order to further improve the user \emph{QoE}, the average quality, the spatial and temporal smoothness of tiles in an FoV are taken into account to refine the quality levels of tiles in an FoV. Assume that there are $M{\rm{ }}\left( {0 < M < N} \right)$  tiles in an FoV. The bitrate and distortion of the $m$-th tile in the FoV of the $l$-th segment are denoted by $\mathbb{R}_{l,FoV,m}^0$ and $\mathbb{D}_{l,FoV,m}^0$, respectively, whereas the bitrate and distortion of the other tiles of the $l$-th segment are denoted by $\mathbb{R}_{l,NFoV,n}^0$ and $\mathbb{D}_{l,NFoV,n}^0$, respectively.

The objective function of the refinement procedure is defined as
\begin{equation}\label{E10}
   {\cal F} = {\theta _1} \cdot {\mathbb{D}_{l,FoV,avg}} + {\theta _2} \cdot {\mathbb{D}_{l,FoV,ss}} + {\theta _3} \cdot {\mathbb{D}_{l,FoV,ts}},
\end{equation}
where ${\theta _1}$, ${\theta _2}$ and ${\theta _3}$ are weighted coefficients of average quality of FoV, spatial smoothness, and temporal smoothness, respectively, which meet the condition: ${\theta _1}{\rm{ + }}{\theta _2}{\rm{ + }}{\theta _3} = 1$. Under extreme conditions, if we only consider the average quality of FoV, the parameters should be set ${\theta _1}=1$, ${\theta _2}=0$ and ${\theta _3}=0$. And if we only consider the spatial and temporal consistency, the parameters could be set ${\theta _1}=0$, ${\theta _2}=0.5$ and ${\theta _3}=0.5$. These three parameters can be set according to the characteristics of the video content. Changing the values of these three parameters will not cause the algorithm to crash. If the video content is complicated and the spatiotemporal texture changes greatly between tiles, we increase the values of ${\theta _2}$ and ${\theta _3}$ to ensure the spatial and temporal consistency, while the video content is simple and consistent, we raise the value of ${\theta _1}$ to ensure the average quality of FoV.

The average quality ${\mathbb{D}_{l,FoV,avg}}$ of the FoV can be calculated by
\begin{equation}\label{E11}
   {\mathbb{D}_{l,FoV,avg}}=\frac{1}{M}\sum\nolimits_{m = 1}^M {\mathbb{D}_{l,FoV,m}^{\cal F}},
\end{equation}
where ${\mathbb{D}_{l,FoV,m}^{\cal F}}$ is the distortion of the $m$-th tile in the FoV that should be determined, the spatial smoothness ${\mathbb{D}_{l,FoV,ss}}$ of the tiles in the FoV is represented by the standard derivation:
\begin{equation}\label{E12}
   {\mathbb{D}_{l,FoV,ss}} = {\left[ {\frac{1}{M}\sum\nolimits_{m = 1}^M {{{\left( {\mathbb{D}_{l,FoV,m}^{\cal F} - {\mathbb{D}_{l,FoV,avg}}} \right)}^2}} } \right]^{1/2}},
\end{equation}
and the temporal smoothness ${\mathbb{D}_{l,FoV,ts}}$ is defined as
\begin{equation}\label{E13}
   {\mathbb{D}_{l,FoV,ts}} = \frac{1}{2}\left| {{\mathbb{D}_{l - 1,FoV,avg}} - {\mathbb{D}_{l,FoV,avg}}} \right|.
\end{equation}

 %'算法伪代码2'
\begin{table}[htbp]
  \centering
    \begin{tabular}{l}
    \toprule
    \textbf{\textit{Algorithm 2}: Coarse to Fine Bit Allocation Algorithm} \\
    \midrule
    Stage 1: Coarse Bit Allocation\\
    \ 1: Confirm the rate distortion model parameters of ${\alpha}$ and ${\beta}$ for all the\\
    \quad\ tiles in the \emph{l}-th segment.\\
    \ 2: Solve (8) by KKT conditions as shown in (9) to obtain ${\mathbb{R}_{l,n}}$.\\
    Stage 2: Fine Bit Allocation\\
        \ 1: Confirm the starting point, $\mathbb{R}_{l,FoV,m}^0$ and $\mathbb{D}_{l,FoV,m}^0$. Note that\\
             \quad \ \ $\mathbb{D}_{l,FoV,m}^0$ corresponds to a quality level ${u_{l,FoV,m}}$.\\
        \ 2: Build a candidate set $\mathcal{A}$ for possible combinations of tiles in FoV,\\
             \quad \ i.e., the initial set ${\cal A} = \left\{ {{\emph{\textbf{a}}_\textbf{1}}} \right\}$, where $ {\emph{\textbf{a}}_\textbf{1}} = \left( {{u_{l,FoV,1}}, \cdots ,} \right.$\\
             \quad \ ${\left. {{u_{l,FoV,M}}} \right)^T}$. Note the initial number of elements of ${\cal A}$ is ${\cal J}=1$,\\
        \ 3: \textbf{set} $j=1$\\
        \ 4: \textbf{while} $j\leq {\cal J}$\\
        \ 5: \quad \textbf{for} $m \leftarrow 1$ \textbf{to} $M$\\
        \ 6: \qquad Confirm the quality levels of ${\textbf{\emph{a}}_\textbf{\emph{j}}} = \left( {{u_{l,FoV,1}}, \cdots ,{u_{l,FoV,m}},} \right. $\\
             \ \quad \qquad $\cdots ,{\left. {{u_{l,FoV,M}}} \right)^T}$, and fix the quality level of all the other\\
             \ \quad \qquad  tiles except for the $m$-th tile\\
        \ 7:\ \qquad \textbf{for} $k \leftarrow 1$ \textbf{to} $U$\\
        \ 8:\ \quad \qquad Change the quality level of the $m$-th tile to be $u_{_{l,FoV,m}}^k$\\
        \ 9:\ \quad \qquad Check the constraints of (14)\\
       10:\ \quad \qquad \textbf{if} (14) holds\\
       11:\qquad \qquad \textbf{if} ${\left( {{u_{l,FoV,1}}, \cdots ,u_{_{l,FoV,m}}^k, \cdots ,{u_{l,FoV,M}}} \right)^T}$ not exists\\
       \ \quad \qquad \qquad  in ${\cal A}$\\
       12:\quad \qquad \qquad ${\textbf{\emph{a}}_{{\cal J}+1}} = \left( {{u_{l,FoV,1}}, \cdots ,u_{_{l,FoV,m}}^k,} \right. \cdots , {\left. {{u_{l,FoV,M}}} \right)^T}$\\
       \ \qquad \qquad \qquad is added to ${\cal A}$,\\
       13:\quad \qquad \qquad ${\cal J} \leftarrow {\cal J}+1$\\
       14:\qquad \qquad \textbf{end if}\\
       15:\ \quad \qquad \textbf{end if}\\
       16:\ \qquad \textbf{end for}\\
       17:\quad \textbf{end for}\\
       18:\quad $j \leftarrow j+1$\\
       19: \textbf{end while}\\
       20: Count the number (denoted by ${\cal J}$) of elements in set ${\cal A}$\\
       21: \textbf{for} $i \leftarrow 1$ \textbf{to} ${\cal J}$\\
       22:\quad Calculate ${{\cal F}_i}$\\
       23:\quad \textbf{if} ${{\cal F}_i} < {{\cal F}_{i - 1}}$\\
       24:\qquad $\mathbb{R}_{l,FoV,m}^{\cal F} = \mathbb{R}_{l,FoV,m}^i$\\
       25:\quad \textbf{end if}\\
       26: \textbf{end for}\\
    \bottomrule
    \end{tabular}
  \label{tab:addlabel}
\end{table}

By constraining the total bitrate and the average distortion of tiles in the FoV to be not larger than the predefined thresholds, i.e., ${\mathbb{R}_{th}}$ and ${\mathbb{D}_{th}}$, the problem of the refinement procedure can be formulated as
\begin{equation}\label{E14}
   \begin{array}{*{20}{c}}
{{\mathop {\min }\limits_{\left( {\mathbb{R}_{l,FoV,m}^{\cal F},\mathbb{D}_{l,FoV,m}^{\cal F}} \right)}{\cal F}},}\vspace{1ex}\\
{s.t.\left\{ \begin{array}{l}
\left| {\sum\nolimits_{m = 1}^M {\mathbb{D}_{l,FoV,m}^{\cal F}}  - \sum\nolimits_{m = 1}^M {\mathbb{D}_{l,FoV,m}^0} } \right| \le {\mathbb{D}_{th}}\vspace{1.5ex}\\
\left| {\sum\nolimits_{m = 1}^M {\mathbb{R}_{l,FoV,m}^{\cal F}}  - \sum\nolimits_{m = 1}^M {\mathbb{R}_{l,FoV,m}^0} } \right| \le {\mathbb{R}_{th}}\vspace{1.5ex}\\
\sum\nolimits_{m = 1}^M {\mathbb{R}_{l,FoV,m}^{\cal F}} + \sum\nolimits_{n = 1}^{N - M} {\mathbb{R}_{l,NFoV,n}^0} \le {\mathbb{R}_{l,request}.}
\end{array} \right.}
\end{array}
\end{equation}

Because the problem in the objective function of Eq. (14) is non-convex, even not analytical, we used a search-based method to solve it, as shown in \textbf{\emph{Algorithm 2}} in which $\mathbb{R}_{l,FoV,m}^0$ and $\mathbb{D}_{l,FoV,m}^0$ are used as the starting point. Fortunately, the search space is not too large ($M$ tiles with $U$ quality levels in the FoV, resulting in ${U^M}$ possible combinations), and the solution can be found effectively and efficiently by the search-based method.

\section{Experimental Results} % IV
In this section, we will first verify the performance of the proposed buffer-quality-based rate adaptation algorithm and then compare the proposed tile-based bit allocation algorithm with the baseline and two state-of-the-art algorithms.
\subsection{Performance of the Buffer-quality-based Rate Adaptation Algorithm} %IV--A
The proposed buffer-quality-based algorithm (denoted as \textbf{BQA}) is compared with \textbf{QFA}[26][27] and \textbf{BFA}[29][30]. As shown in Fig. 7, two kinds of network environments, i.e., stable channel with a fixed bandwidth of 10Mbps and Markovian channel with a transition probability of $p_{t}$=0.5 [27], were simulated. In the proposed \textbf{BQA}, ${b_{min}}$ and ${b_{max}}$ are set to be 10 seconds and 12 seconds, respectively. As shown in Fig. 8, because \textbf{QFA} will request the bitrates that match (less than or equal to) the bandwidth, its buffer length is smaller than \textbf{BFA} and \textbf{BQA}. For the \textbf{BFA}, the buffer length is kept at a relatively high level, but the downloaded segment bitrate fluctuates seriously, which is definitely not good for the user's experience. Besides, the buffer length is not stable. At the same time, we can also see that the proposed \textbf{BQA} makes a compromise between the \textbf{QFA} and \textbf{BFA}. The bitrates of the downloaded video segments are comparatively stable, while the buffer variations are also controlled at the predefined range (from 10 seconds to 12 seconds).

Based on the proposed \textbf{BQA}, we will further compare the tile-based bit allocation algorithms.

\subsection{Comparison of the Tile-based Bit Allocation Algorithms} %IV--B
Three 360-degree video sequences \emph{AerialCity} (with camera motion), \emph{DrivingInCountry} (with camera motion), and \emph{PoleVault} (without camera motion), with resolution of 3840$\times$1920 are used as test sequences. The tested video sequences are spatially partitioned into 4 rows and 6 columns, i.e., 24 tiles. Each tile is then temporally divided into multiple segments with a fixed display time of 2 seconds. Afterwards, each tile is encoded with 16 quality levels (i.e., $U$=16) by H.265/HEVC test model version 16.14 (HM.16.14). The corresponding bitrates are \{150kbps, 300kbps, 450kbps, 600kbps, 750kbps, 900kbps, 1050kbps, 1200kbps, 1350kbps, 1500kbps, 1650kbps, 1800kbps, 1950kbps, 2100kbps, 2250kbps and 2400kbps\}. Because the display time of the tested sequences is only 10 seconds (5 segments), we loop the display in the experiments.

\begin{figure}
\centering
\subfigure[]{
\label{fig7:subfig:a} %% label for first subfigure
\includegraphics[width=4.25cm]{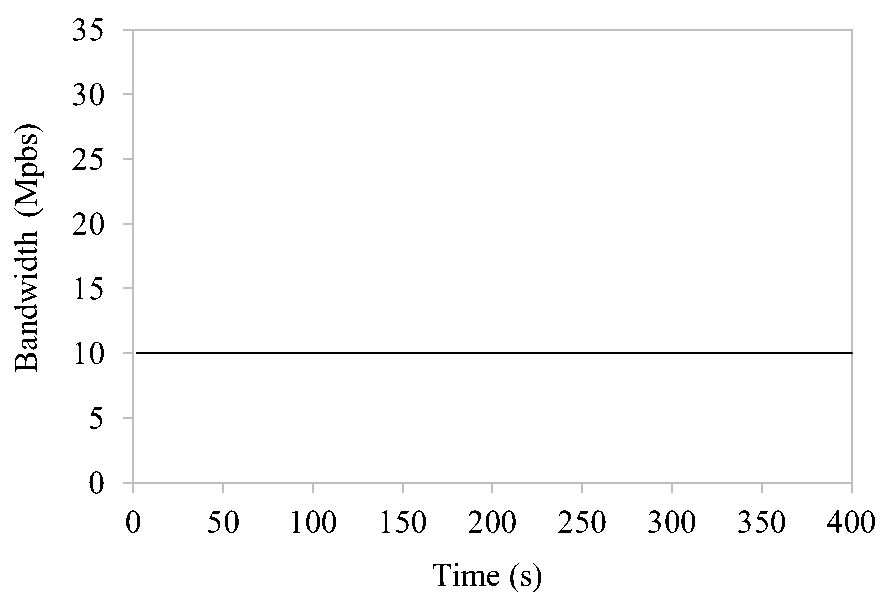}}
\subfigure[]{
\label{fig7:subfig:b} %% label for second subfigure
\includegraphics[width=4.25cm]{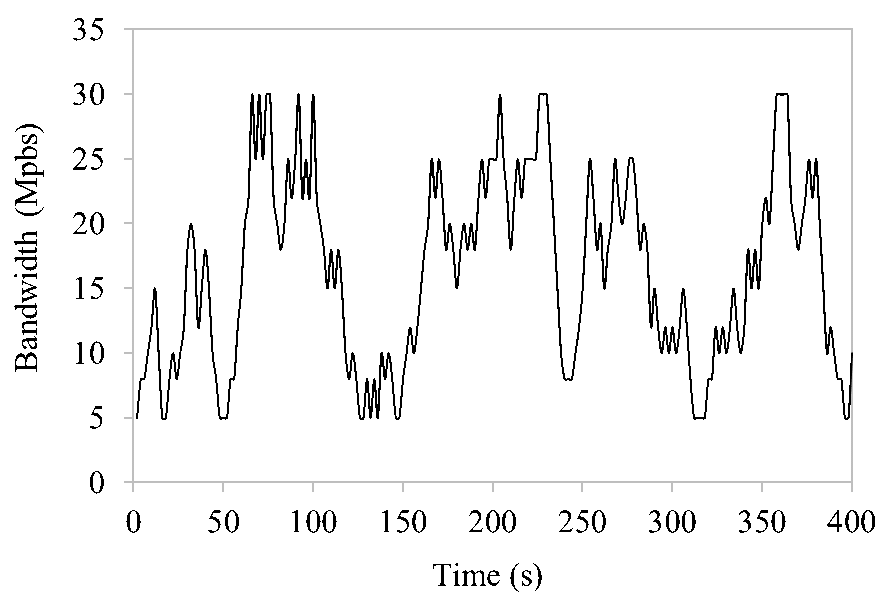}}
\caption{Two kinds of network channel environments, (a) fixed bandwidth of 10Mbps, (b) Markovian channel.}
\label{fig7}
\end{figure}

\begin{figure}
\centering
\subfigure[]{
\label{fig8:subfig:a}
\includegraphics[width=4.25cm]{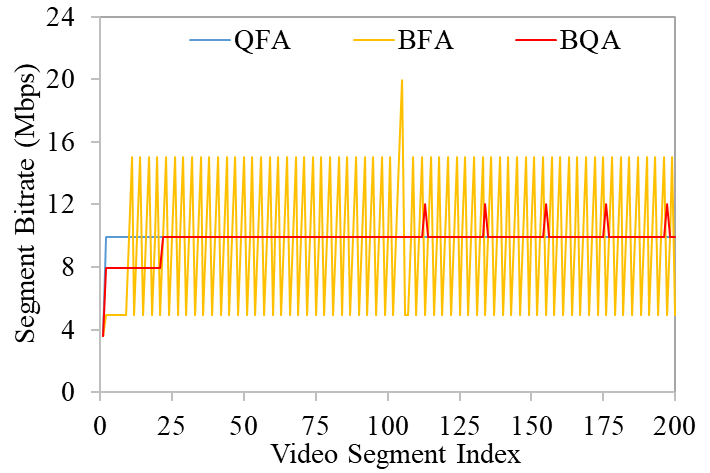}}
\subfigure[]{
\label{fig8:subfig:b}
\includegraphics[width=4.25cm]{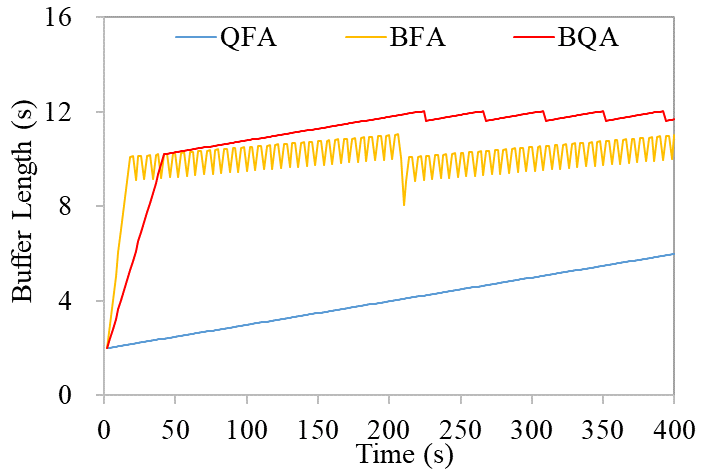}}
\subfigure[]{
\label{fig8:subfig:c}
\includegraphics[width=4.25cm]{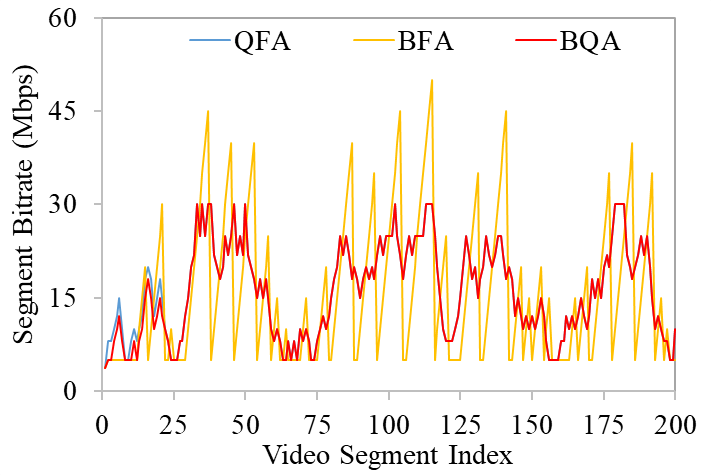}}
\subfigure[]{
\label{fig8:subfig:d}
\includegraphics[width=4.25cm]{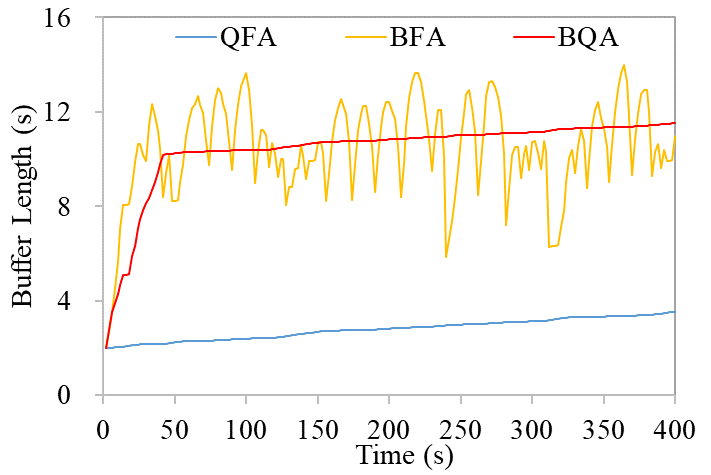}}
\caption{Simulation results, (a) requested segment bitrates under fixed 10Mbps bandwidth channel, (b) buffer length under fixed 10Mbps bandwidth channel, (c) requested segment bitrates under Markovian channel, (b) buffer length under Markovian channel.}
\label{fig8}
\end{figure}

Note that the rate-distortion model parameters ${\alpha _{l,n}}$ and ${\beta _{l,n}}$ of the $n$-th tile in the $l$-th segment in Eq.(7) can be calculated by statistical regression based on the encoding results, as shown in Fig. 9, in which ``r1c1", ``r1c2", etc., denote the corresponding tiles located at row 1 and column 1, row 1 and column 2, etc. We can see that the model parameters of tiles are quite different. Therefore, in order to improve user \emph{QoE}, it is not enough to only consider the bitrates.

\begin{figure}
\centering
\subfigure[]{
\label{fig9:subfig:a}
\includegraphics[width=4.25cm]{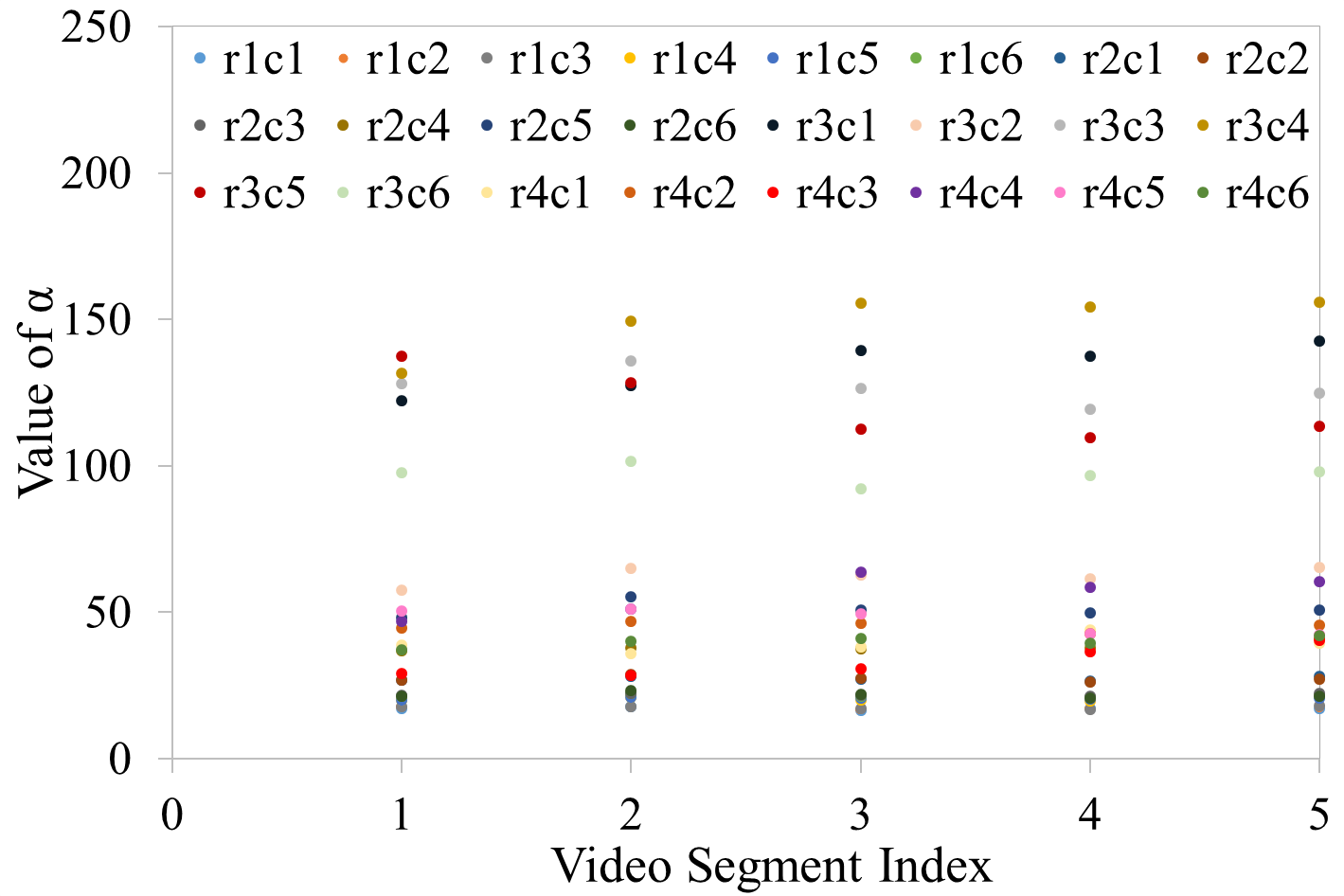}}
\subfigure[]{
\label{fig9:subfig:b}
\includegraphics[width=4.25cm]{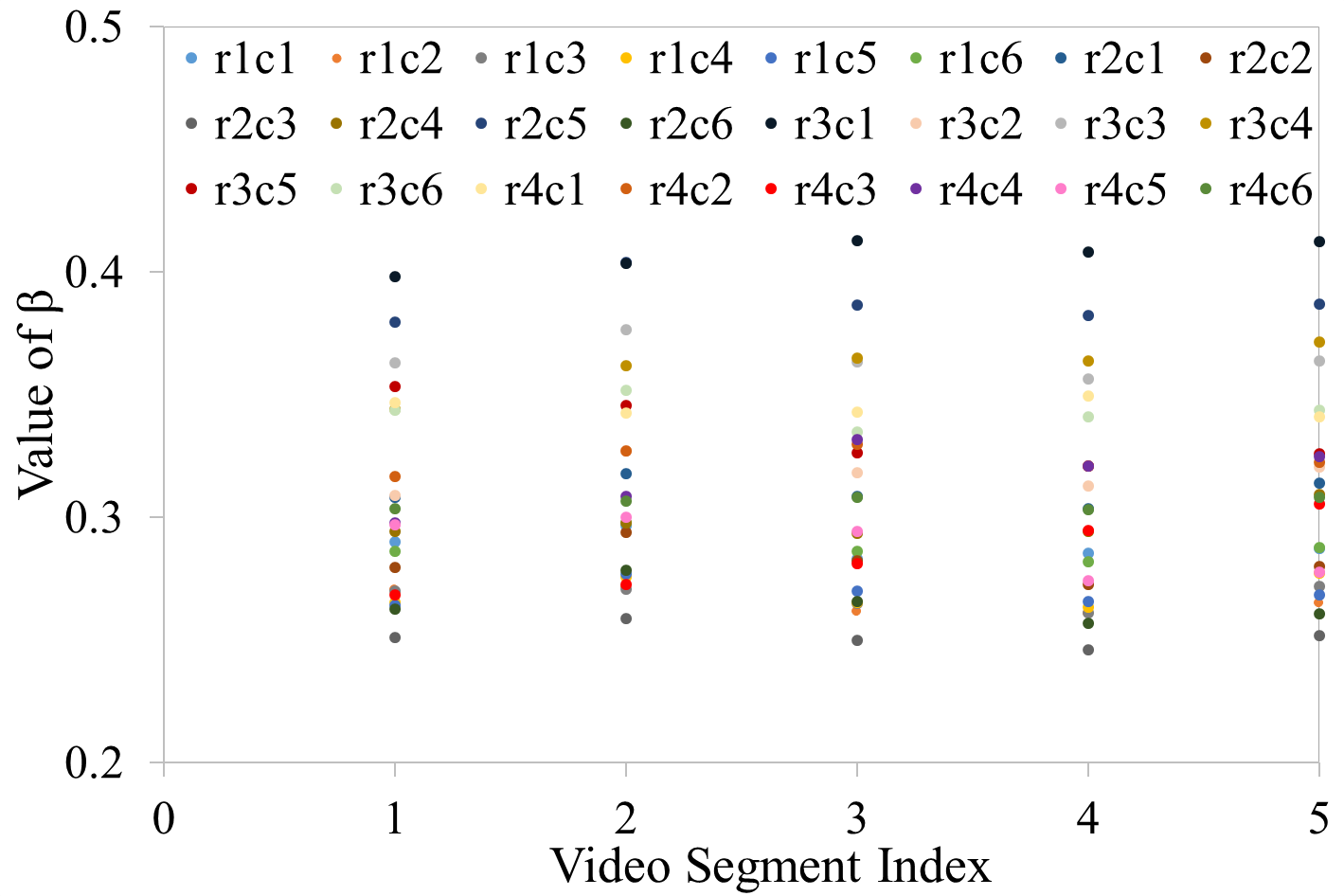}}
\caption{Rate-distortion model parameters (a) ${\alpha _{l,n}}$, (b)${\beta _{l,n}}$ of the $n$-th tile in the $l$-th segment.}
\label{fig9}
\end{figure}

According to the random walk theory [52], the initial viewport is selected randomly. Then the Gaussian model is used to predict the viewport at the beginning of each segment with $\mu$=11 and ${\sigma ^2}$=4 for video sequences with camera motion (i.e. \emph{AerialCity} and \emph{DrivingInCountry}) and $\mu$=11 and ${\sigma ^2}$=9 for video sequences without camera motion (i.e. \emph{PoleVault}). After that, the Zipf model is used to determine the priority of each tile in a segment. For the spatial and temporal smoothness in Eq. (10), we set ${\theta _1}$=0.2, ${\theta _2}$=0.3 and ${\theta _3}$=0.5 empirically. Besides, ${\mathbb{D}_{th}}$ and ${\mathbb{R}_{th}}$ for solving (14) are set to be 0.4 and 2Mbps, respectively.

To demonstrate the advantage of our proposed bit allocation method, the average allocation method (denoted by \textbf{AA Method}) is used as the baseline. The \textbf{AA Method} allocates the available bitrates to all the tiles equally. Besides, the state-of-the-art algorithms including adaptive allocation method [38] (denoted by \textbf{AdapA Method}) and partial delivery method [20](denoted by \textbf{PD Method}), are also compared. In the \textbf{AdapA Method}, the requested bitrate will first be allocated to the tiles with relatively high priorities (e.g., tiles in FoV). Note that some tiles with low priorities may not be downloaded depending on the requested bitrate, the current view angle, and the view switching probability. The \textbf{PD Method} only requests the tiles in the FoV with the highest possible quality representation and all other tiles are not requested at all. Besides, we also compared the performance of the first stage of the proposed method (denoted by \textbf{Proposed Method w/o ST}), i.e., the solution of Eq. (6) in which the spatial and temporal smoothness are not considered. To ensure the fairness in the experiments, the \textbf{BQA} is used to determine the requested video bitrates for all the compared bit allocation methods, and the buffer thresholds $b_{min}$ and $b_{max}$ are set as 10 seconds and 20 seconds in the experiments.

The performance is evaluated by the following quotas:

(a) actual downloaded bitrates of tiles in a segment (denoted by \emph{Actual Bitrate});

(b) weighted PSNR of all tiles in a video segment (denoted by \emph{Weighted PSNR},i.e.,
\begin{equation}\label{E15}
{\emph{Weighted PSNR} = }\sum\nolimits_{n = 1}^N {\left[ {{p_{l,n}} \times 10{{\log }_{10}}\left( {\frac{{{{255}^2}}}{{{\mathbb{D}_{l,n}}}}} \right)} \right]},
%{\rm{Weighted PSNR = }}\sum\nolimits_{n = 1}^N {\left[ {{p_{l,n}} \times 10{{\log }_{10}}\left( {\frac{{{{255}^2}}}{{{\mathbb{D}_{l,n}}}}} \right)} \right]},
\end{equation}
where $p_{l,n}$ and $\mathbb{D}_{l,n}$ represent the priority and the distortion of the \emph{n}-th tile in the \emph{l}-th video segment;

(c) actual downloaded bitrates of tiles in FoV (denoted by \emph{FoV Actual Bitrate});

(d) average PSNR of tiles in FoV, i.e., (denoted by \emph{FoV Average PSNR});

(e) standard deviation of PSNR of tiles in FoV (denoted by \emph{FoV PSNR Std});

(f) average PSNR difference of FoVs between two consecutive segments (denoted by \emph{FoV PSNR Temporal Difference});

(g) \emph{Buffer length};

(h) $\mathcal{F}$ \emph{value}, i.e., the objection value of Eq. (10).

We then established a test platform for video delivery based on the guidelines of DASH Industry Form [53]. The test platform consists of two parts: an \emph{Apache} HTTP webserver and a user client. The proposed algorithm is validated under two network environments:

\emph{Case1: the network connection between the webserver and the user client is controlled by DummyNet} [54];

\emph{Case 2: the server and the user client are connected by the actual campus wireless network of Shandong University}.

For \emph{Case 1}, the staged throughput variation [55], as shown in Fig.10, is monitored periodically. We set $L_{0}$ in Eq. (1) to be 1, i.e., the current bandwidth is predicted by download time and bitrate of the previous segment. For \emph{Case 2}, because of the frequent fluctuations in the actual network throughput, the user client is more likely to suffer from buffer starvation. To tackle this problem, we set $L_{0}$=4. Besides, we conduct the experiments for 10 times, and the average results are reported.\vspace{1ex}
\begin{figure}
\centering
\includegraphics[width=6cm]{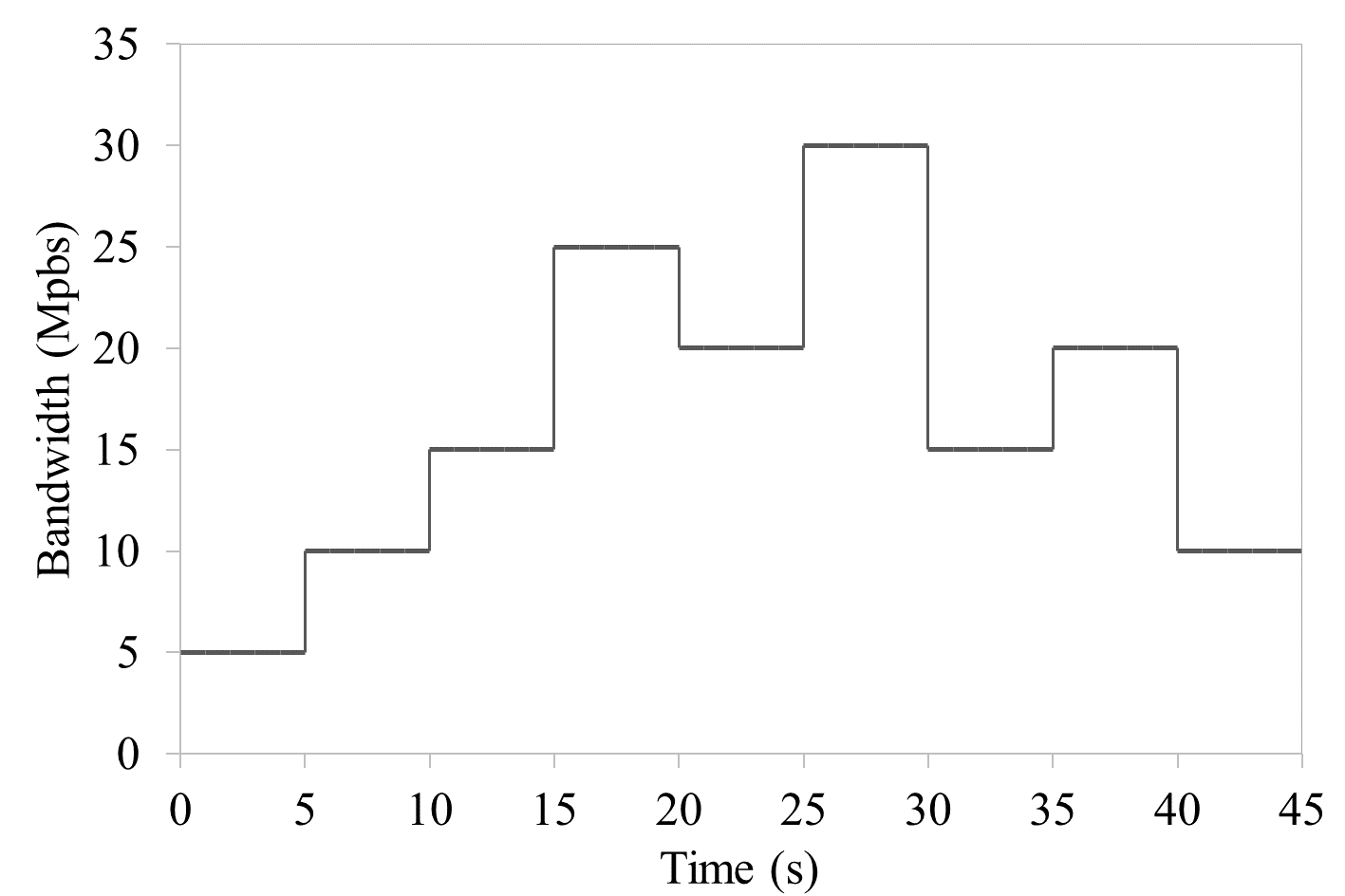}
\caption{Server bandwidth with staged variation.}
\label{fig10}
\end{figure}

\noindent \emph{(i) Results of Case 1} \vspace{1ex}%IV--B--1

Because \emph{Dummynet} cannot control the network throughput steadily, the actual channel throughput fluctuates irregularly based on the curve of Fig. 10. Therefore, even when the bandwidth is fixed, the actual bitrates will also fluctuate.

Detailed experimental results of the three video sequences are shown in Figs.11-13. Taking \emph{AerialCity} as an example, from Fig.11(a), we can see that the \emph{Actual Bitrates} of the five methods are similar. When evaluating the \emph{FoV Actual Bitrate}, as shown in Fig. 11(b), the \emph{FoV Actual Bitrates} of the \textbf{AdpaA Method} and the \textbf{PD Method} are obviously larger than the others because the requested bitrate will be first allocated to tiles in an FoV, while the remaining tiles are not be downloaded. In addition, the \emph{FoV Actual Bitrate} of the \textbf{AA method} is the lowest.

Figs.11(c) and (d) compare the \emph{Weighted PSNR} and \emph{FoV Average PSNR} of each segment, respectively. The \emph{Weighted PSNRs} of the \textbf{AdpaA Method} and the \textbf{PD Method} are obviously lower than the others because some tiles that do not belong to FoVs are not downloaded. Nevertheless, the \emph{FoV Average PSNR} of the \textbf{PD Method} is higher because it prioritizes the tiles in an FoV. As expected, the \textbf{AA Method} has the lowest \emph{FoV Average PSNR}, and the values of the \textbf{Proposed Method w/o ST} and the \textbf{Proposed Method} are in between with the \textbf{PD Method} and the \textbf{AA Method}.

%图11
\begin{figure*} %加“*”表示不分栏
\centering
\subfigure[]{
\label{fig11:subfig:a}
\includegraphics[width=4cm]{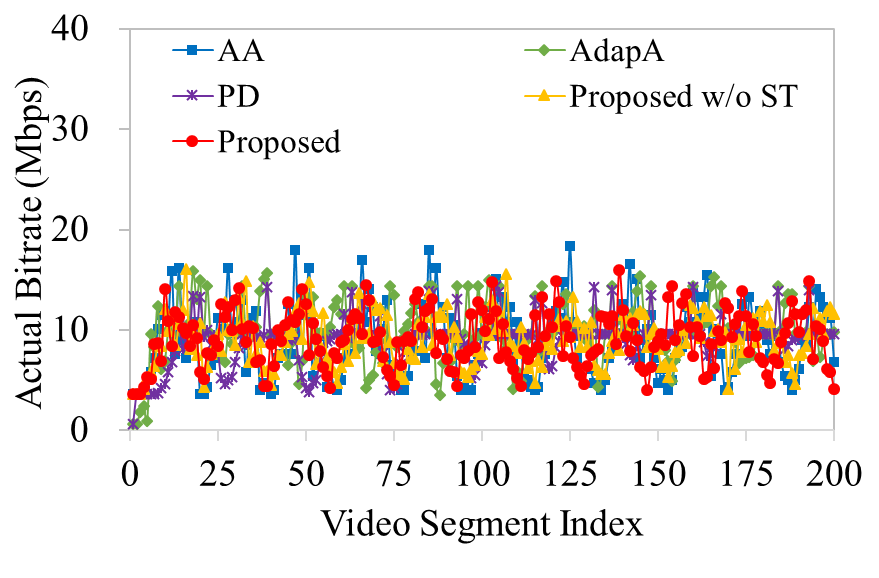}}
\subfigure[]{
\label{fig11:subfig:b}
\includegraphics[width=4cm]{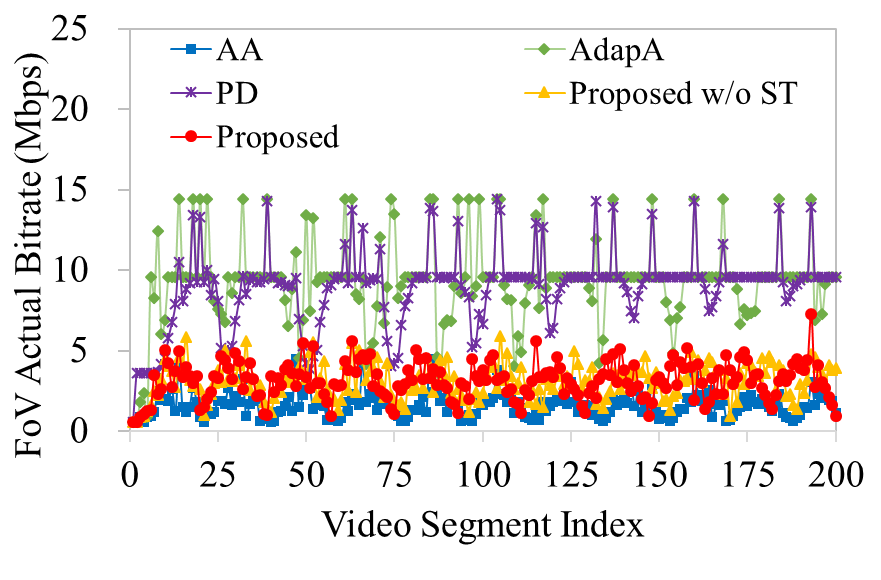}}
\subfigure[]{
\label{fig11:subfig:c}
\includegraphics[width=4cm]{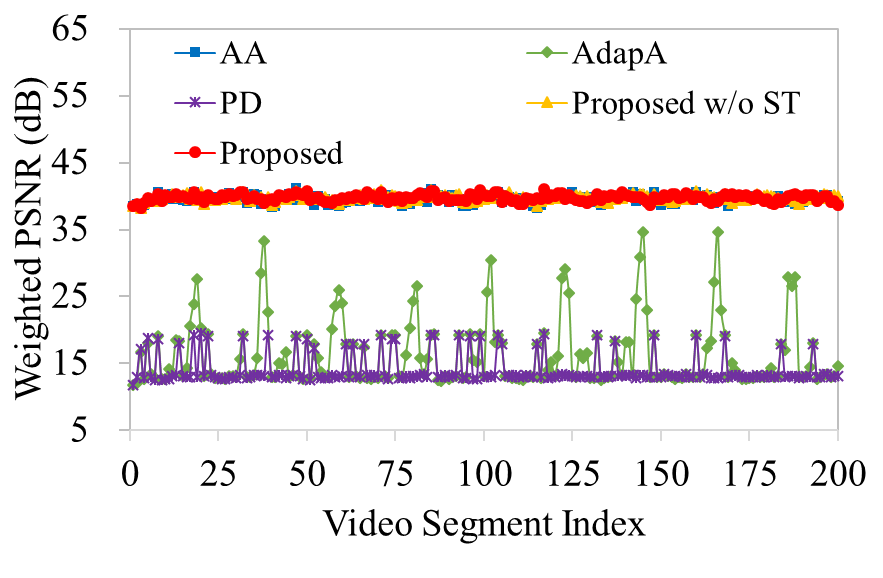}}
\subfigure[]{
\label{fig11:subfig:d}
\includegraphics[width=4cm]{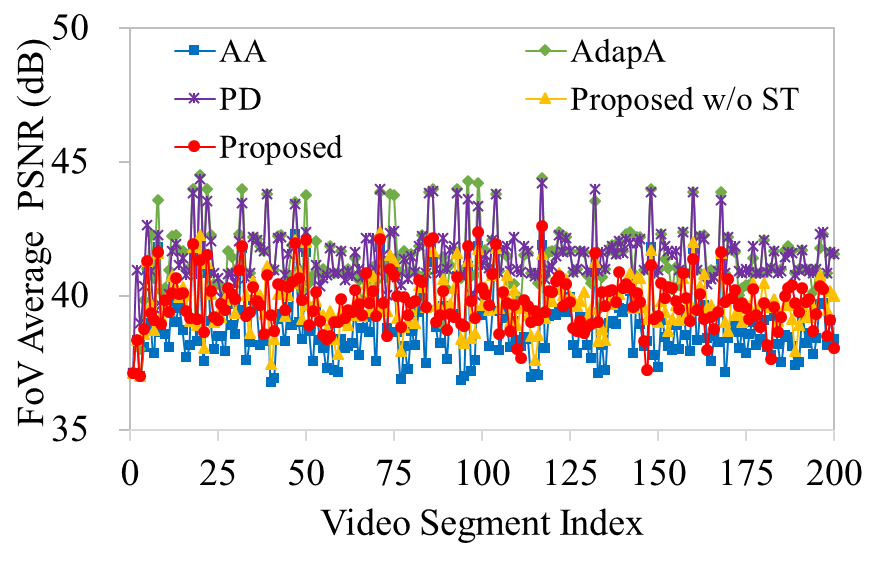}}
\subfigure[]{
\label{fig11:subfig:e}
\includegraphics[width=4cm]{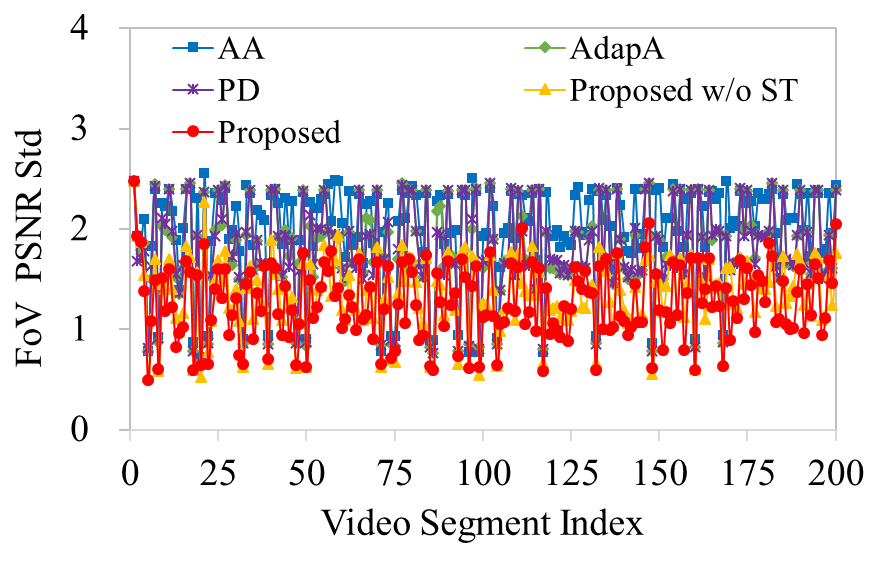}}
\subfigure[]{
\label{fig11:subfig:f}
\includegraphics[width=4cm]{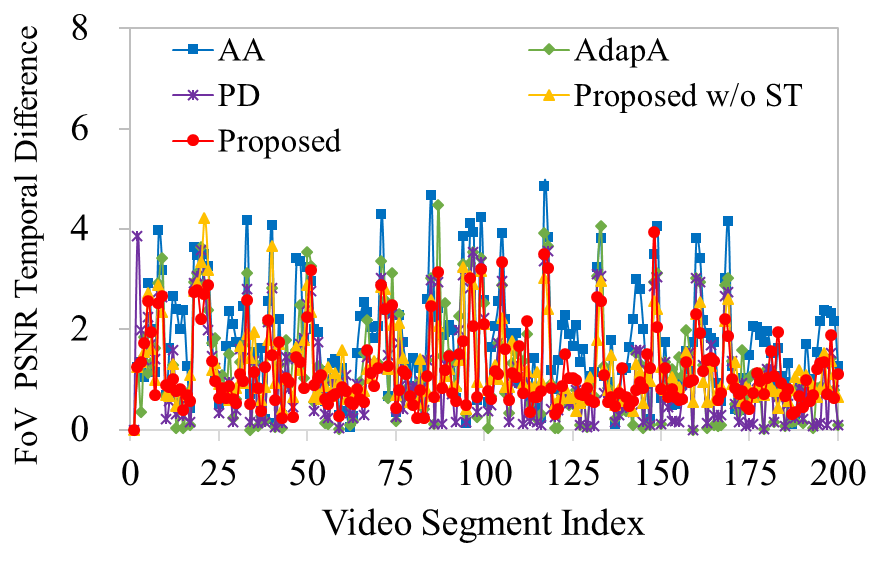}}
\subfigure[]{
\label{fig11:subfig:g}
\includegraphics[width=4cm]{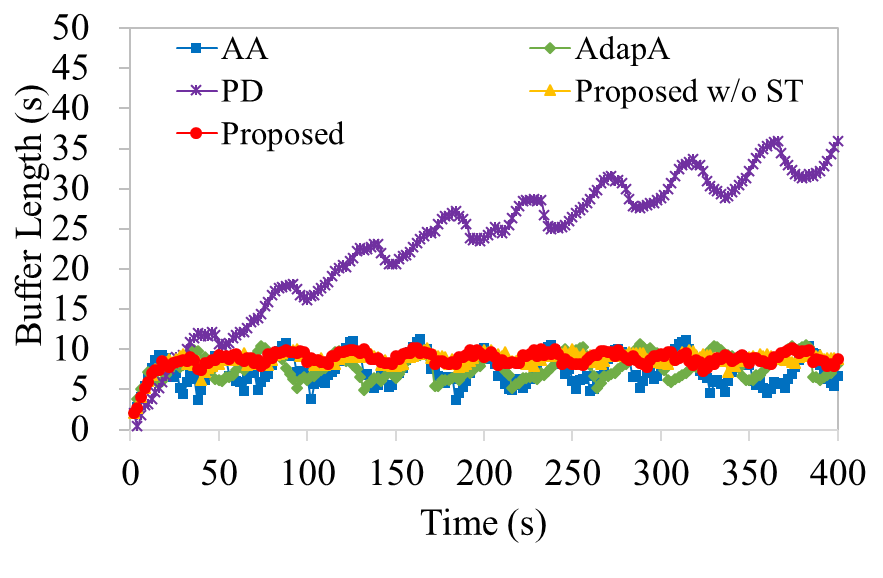}}
\subfigure[]{
\label{fig11:subfig:h}
\includegraphics[width=4cm]{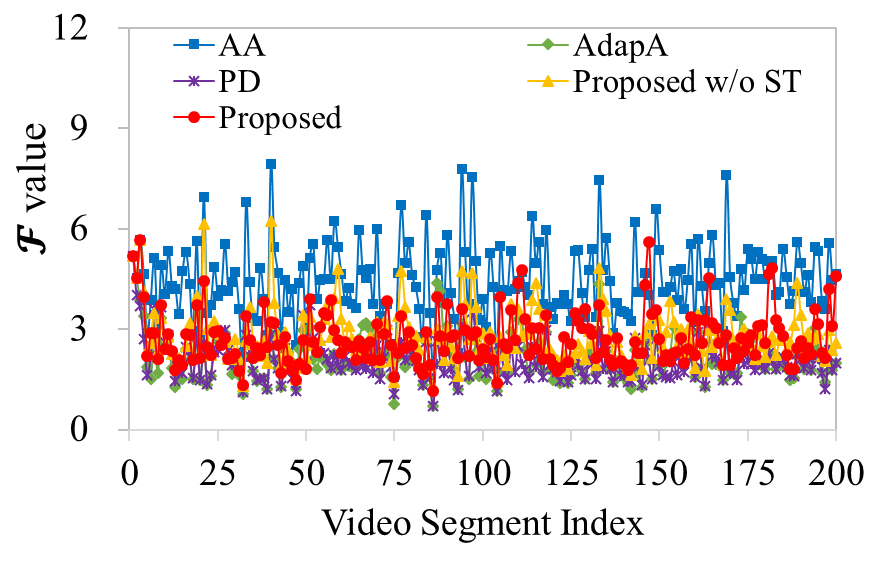}}
\caption{Results of \emph{Case 1} with video sequence \emph{AerialCity}.}
\label{fig11}
\end{figure*}

%表1    QUANTITATIVE COMPARISIONS OF CASE 1 WITH VIDEO SEQUENCE AERIALCITY
% Table generated by Excel2LaTeX from sheet 'AerialCity-虚-制表-1.2'
\begin{table*}[htbp]
 \scriptsize
  \centering
  \caption{QUANTITATIVE COMPARISIONS OF CASE 1 WITH VIDEO SEQUENCE AERIALCITY}
    \begin{tabular}{cccccccc}
    \toprule
    \toprule
    Sudden view switching &       & \textit{FoV Actual} & \textit{FoV} & \textit{FoV} & \textit{FoV  PSNR} &       &  \\
    probability in the duration & Method & \textit{Bitrate} & \textit{Average} & \textit{PSNR Std} & \textit{Temporal} & \textit{$\mathcal{F}$ value} & \textit{QoE} \\
    of each video segment &       & \textit{(Mbps)} & \textit{PSNR (dB)} &       & \textit{Difference} &       &  \\
    \midrule
    \multirow{5}[2]{*}{0\%} & AA    & 1.61  & 38.82  & 2.01  & 1.75  & 4.4533  & 3422.84  \\
          & AdapA & \textbf{9.34 } & 41.42  & 1.80  & 1.20  & 2.1424  & 5745.09  \\
          & PD    & 8.94  & \textbf{41.44 } & 1.80  & \textbf{1.08 } & \textbf{2.0610 } & \textbf{6740.97 } \\
          & Proposed w/o ST & 3.10  & 39.78  & 1.35  & 1.23  & 2.7815  & 5663.52  \\
          & Proposed & 3.14  & 39.76  & \textbf{1.26 } & 1.17  & 2.7216  & 5757.89  \\
    \midrule
    \multirow{5}[2]{*}{5\%} & AA    & 1.63  & 38.87  & 1.99  & 1.77  & 4.4177  & 3414.86  \\
          & AdapA & \textbf{9.16 } & \textbf{40.63 } & 1.89  & 1.99  & 636.1096  & 4643.83  \\
          & PD    & 8.74  & 40.55  & 1.94  & 1.95  & 730.1807  & 5512.03  \\
          & Proposed w/o ST & 3.06  & 39.79  & 1.34  & 1.23  & 2.7745  & 5657.39  \\
          & Proposed & 3.10  & 39.78  & \textbf{1.25 } & \textbf{1.18 } & \textbf{2.7147 } & \textbf{5749.24 } \\
    \midrule
    \multirow{5}[2]{*}{10\%} & AA    & 1.64  & 38.90  & 1.97  & 1.76  & 4.3906  & 3422.88  \\
          & AdapA & \textbf{8.94 } & 39.60  & 1.95  & 3.00  & 1415.6537  & 3228.17  \\
          & PD    & 8.53  & 39.46  & 1.90  & 3.03  & 1506.5815  & 4005.35  \\
          & Proposed w/o ST & 3.02  & \textbf{39.78 } & 1.33  & 1.23  & 2.7779  & 5655.92  \\
          & Proposed & 3.06  & 39.77  & \textbf{1.25 } & \textbf{1.19 } & \textbf{2.7230 } & \textbf{5739.96 } \\
    \midrule
    \multirow{5}[2]{*}{20\%} & AA    & 1.65  & 38.90  & 1.96  & 1.77  & 4.3836  & 3411.31  \\
          & AdapA & \textbf{8.50 } & 37.80  & 2.13  & 4.69  & 2811.7930  & 836.50  \\
          & PD    & 8.12  & 37.63  & 1.99  & 4.75  & 2888.1290  & 1563.43  \\
          & Proposed w/o ST & 2.96  & \textbf{39.74 } & 1.32  & 1.22  & 2.7882  & 5660.95  \\
          & Proposed & 2.99  & 39.73  & \textbf{1.25 } & \textbf{1.18 } & \textbf{2.7399 } & \textbf{5736.80 } \\
    \bottomrule
    \bottomrule
    \end{tabular}%
  \label{tab:addlabel}%
\end{table*}%

%图12
\begin{figure*} %加“*”表示不分栏
\centering
\subfigure[]{
\label{fig12:subfig:a}
\includegraphics[width=4cm]{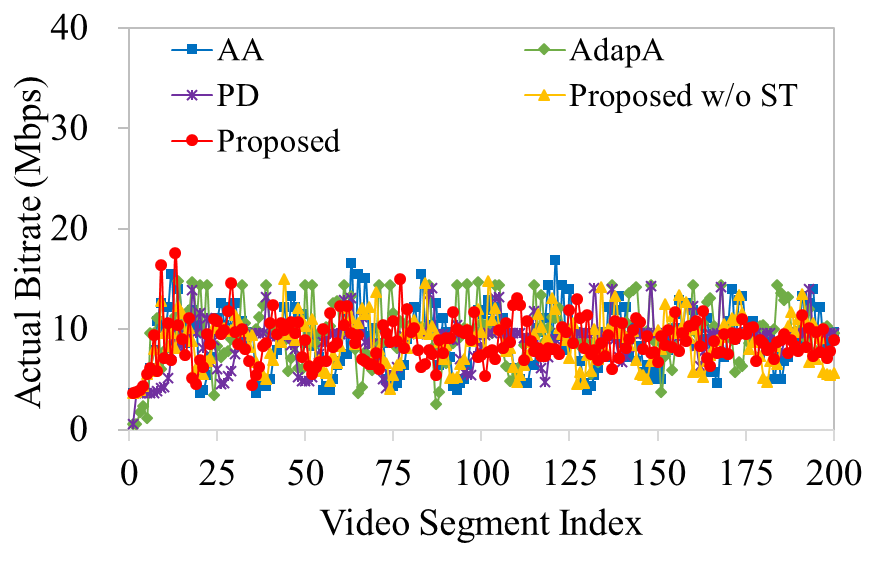}}
\subfigure[]{
\label{fig12:subfig:b}
\includegraphics[width=4cm]{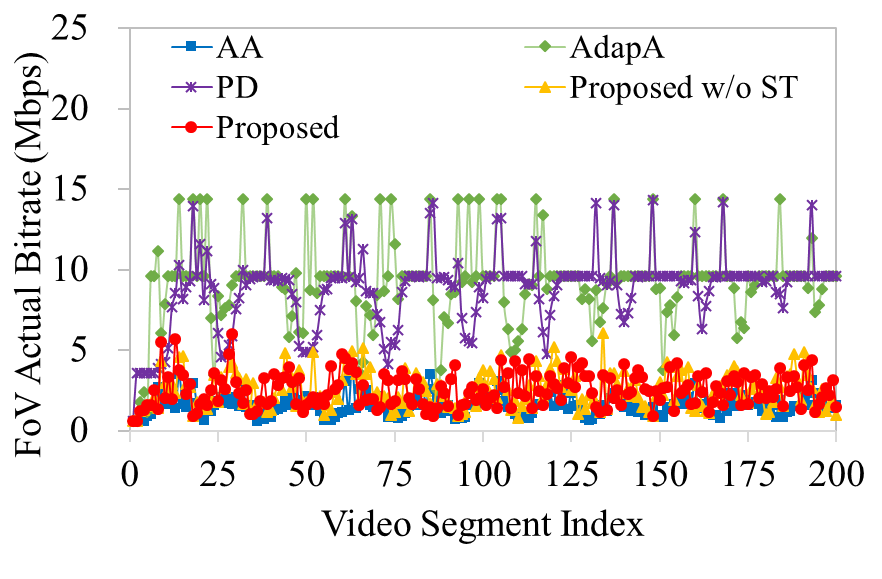}}
\subfigure[]{
\label{fig12:subfig:c}
\includegraphics[width=4cm]{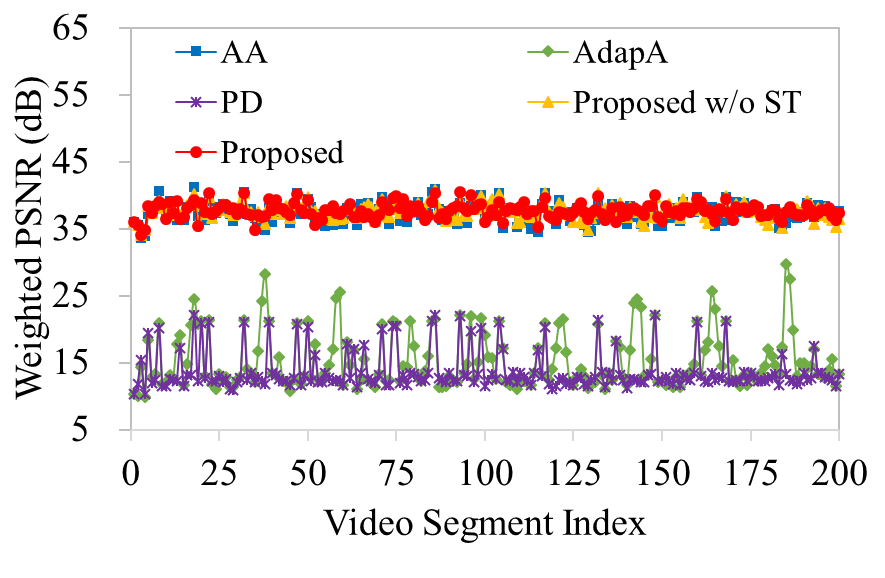}}
\subfigure[]{
\label{fig12:subfig:d}
\includegraphics[width=4cm]{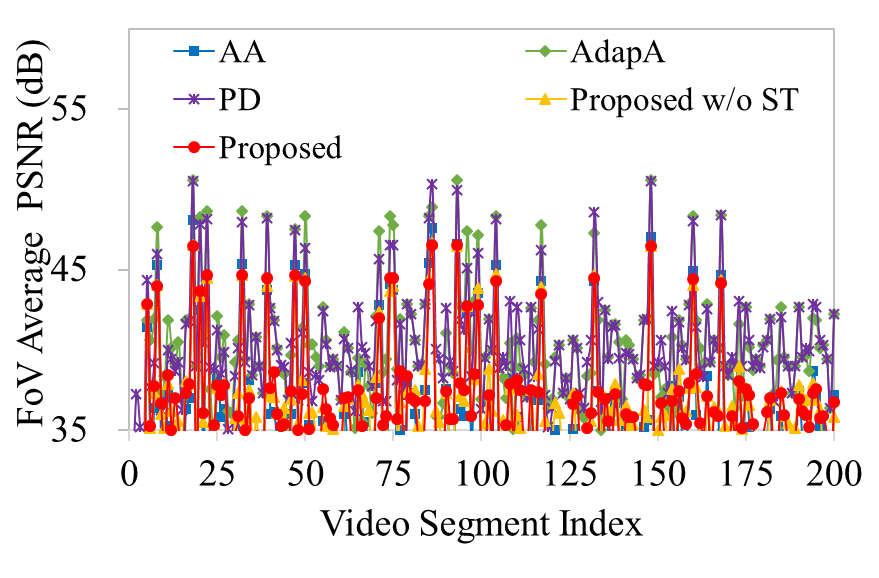}}
\subfigure[]{
\label{fig12:subfig:e}
\includegraphics[width=4cm]{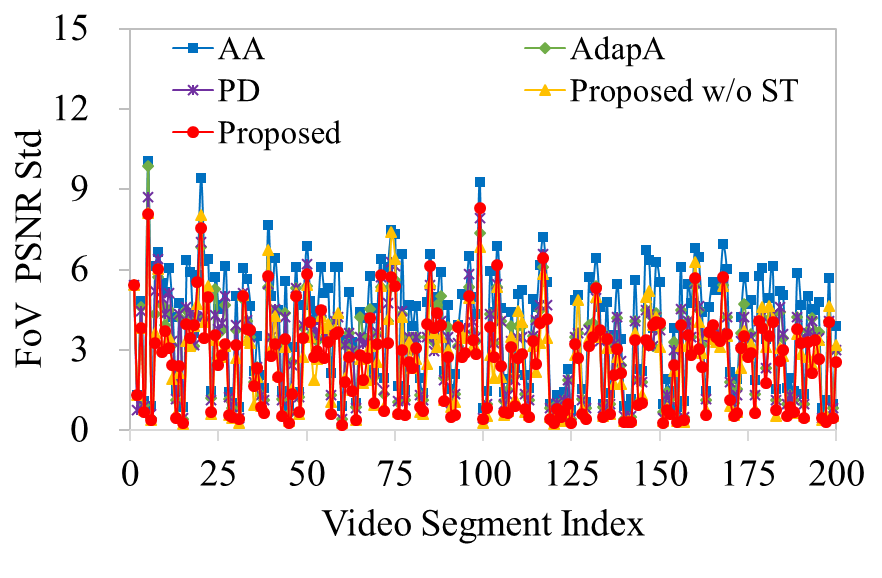}}
\subfigure[]{
\label{fig12:subfig:f}
\includegraphics[width=4cm]{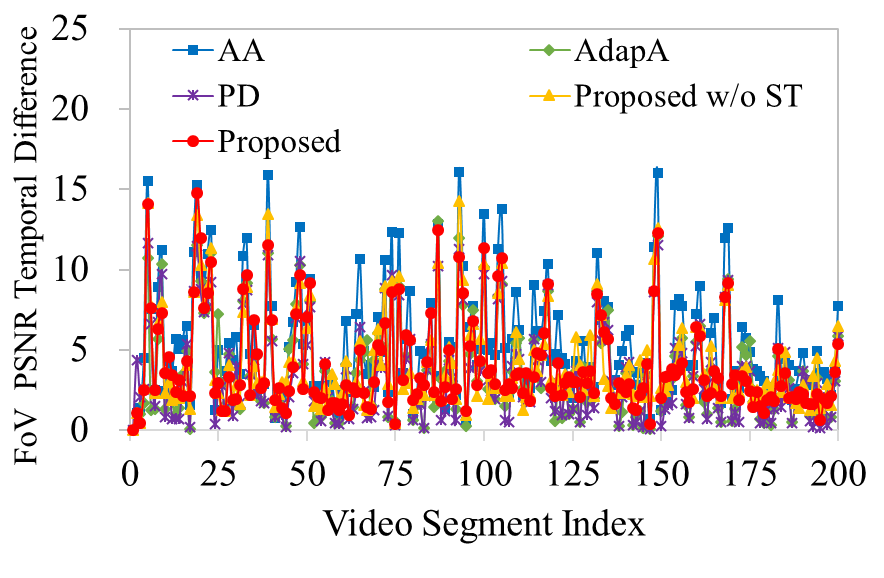}}
\subfigure[]{
\label{fig12:subfig:g}
\includegraphics[width=4cm]{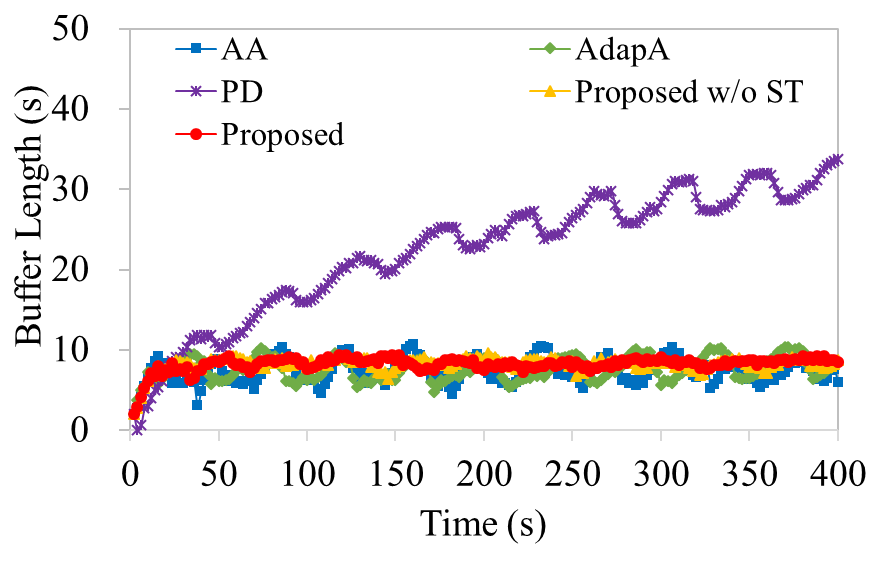}}
\subfigure[]{
\label{fig12:subfig:h}
\includegraphics[width=4cm]{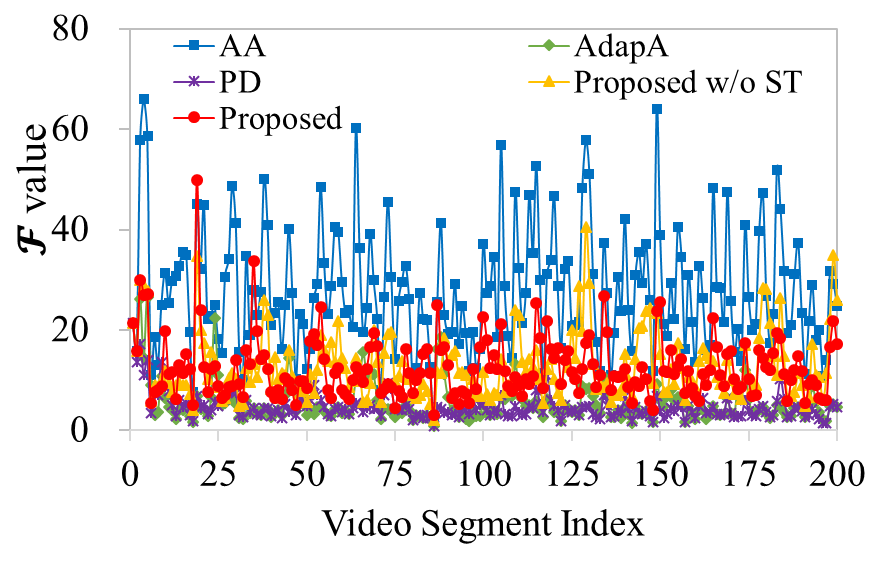}}
\caption{Results of \emph{Case 1} with video sequence \emph{DrivingInCountry}.}
\label{fig12}
\end{figure*}

%表2  QUANTITATIVE COMPARISIONS OF CASE 1 WITH VIDEO SEQUENCE DRIVINGINCOUNTRY
% Table generated by Excel2LaTeX from sheet 'DrivingInCountry- 虚-制表-1.2'
\begin{table*}[htbp]
 \scriptsize
  \centering
  \caption{QUANTITATIVE COMPARISIONS OF CASE 1 WITH VIDEO SEQUENCE DRIVINGINCOUNTRY}
   \begin{tabular}{cccccccc}
    \toprule
    \toprule
    Sudden view switching  &       & \textit{FoV Actual} & \textit{FoV} & \textit{FoV} & \textit{FoV  PSNR} &       &  \\
    probability in the duration & Method & \textit{Bitrate} & \textit{Average} & \textit{PSNR Std} & \textit{Temporal} & \textit{$\mathcal{F}$ value} & \textit{QoE} \\
    of each video segment &       & \textit{(Mbps)} & \textit{PSNR (dB)} &       & \textit{Difference} &       &  \\
    \midrule
       \multirow{5}[2]{*}{0\%} & AA    & 1.61  & 34.85  & 4.03  & 5.32  & 27.7687  & 2234.94  \\
          & AdapA & \textbf{9.22 } & 40.31  & 3.07  & 3.68  & 5.4414  & 6519.90  \\
          & PD    & 8.79  & \textbf{40.38 } & 3.08  & \textbf{3.43 } & \textbf{4.6455 } & \textbf{7645.58 } \\
          & Proposed w/o ST & 2.49  & 36.27  & 2.61  & 4.07  & 12.7353  & 5429.34  \\
          & Proposed & 2.53  & 36.33  & \textbf{2.59 } & 3.95  & 12.4203  & 5592.72  \\
    \midrule
    \multirow{5}[2]{*}{5\%} & AA    & 1.63  & 35.05  & 4.09  & 5.39  & 27.5176  & 2197.93  \\
          & AdapA & \textbf{9.03 } & 39.51  & 3.19  & 4.44  & 710.5147  & 5451.16  \\
          & PD    & 8.60  & \textbf{39.54 } & 3.20  & 4.21  & 732.6536  & \textbf{6540.67 } \\
          & Proposed w/o ST & 2.45  & 36.41  & 2.69  & 4.14  & 12.7793  & 5373.25  \\
          & Proposed & 2.50  & 36.47  & \textbf{2.66 } & \textbf{3.99 } & \textbf{12.3947 } & 5571.77  \\
    \midrule
    \multirow{5}[2]{*}{10\%} & AA    & 1.64  & 35.17  & 4.12  & 5.43  & 27.5912  & 2169.94  \\
          & AdapA & \textbf{8.82 } & \textbf{38.57 } & 3.18  & 5.37  & 1448.0600  & 4148.80  \\
          & PD    & 8.37  & 38.52  & 3.16  & 5.24  & 1508.9930  & 5108.33  \\
          & Proposed w/o ST & 2.44  & 36.49  & 2.73  & 4.19  & 12.9401  & 5337.29  \\
          & Proposed & 2.47  & 36.53  & \textbf{2.70 } & \textbf{4.07 } & \textbf{12.6049 } & \textbf{5494.45 } \\
    \midrule
    \multirow{5}[2]{*}{20\%} & AA    & 1.65  & 34.96  & 4.07  & 5.33  & 27.9048  & 2250.22  \\
          & AdapA & \textbf{8.39 } & \textbf{36.66 } & 3.22  & 6.82  & 2844.7701  & 2022.67  \\
          & PD    & 7.99  & 36.62  & 3.14  & 6.68  & 2890.3337  & 2991.32  \\
          & Proposed w/o ST & 2.44  & 36.28  & 2.70  & 4.13  & 13.2384  & 5369.88  \\
          & Proposed & 2.48  & 36.33  & \textbf{2.68 } & \textbf{4.01 } & \textbf{12.8806 } & \textbf{5527.01 } \\
    \bottomrule
    \bottomrule
    \end{tabular}%
  \label{tab:addlabel}%
\end{table*}%

%图13
\begin{figure*} %加“*”表示不分栏
\centering
\subfigure[]{
\label{fig13:subfig:a}
\includegraphics[width=4cm]{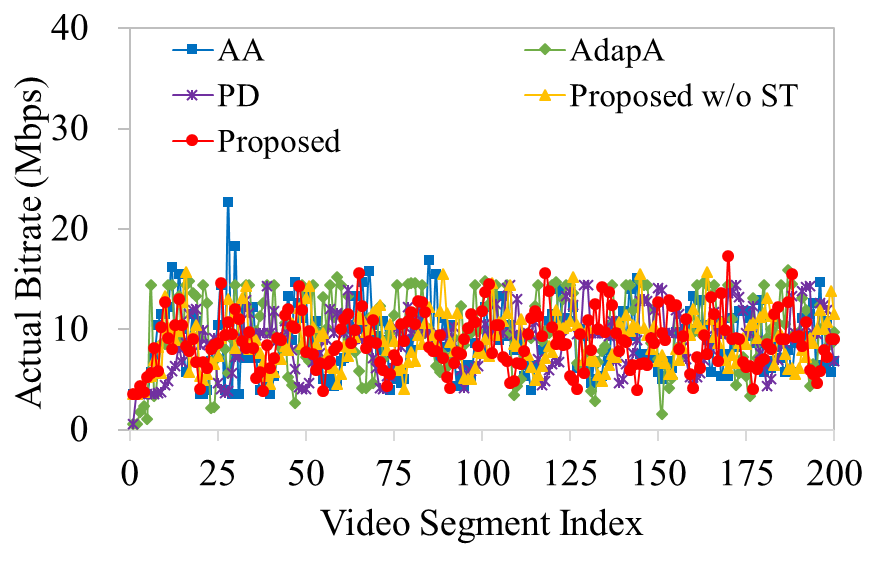}}
\subfigure[]{
\label{fig13:subfig:b}
\includegraphics[width=4cm]{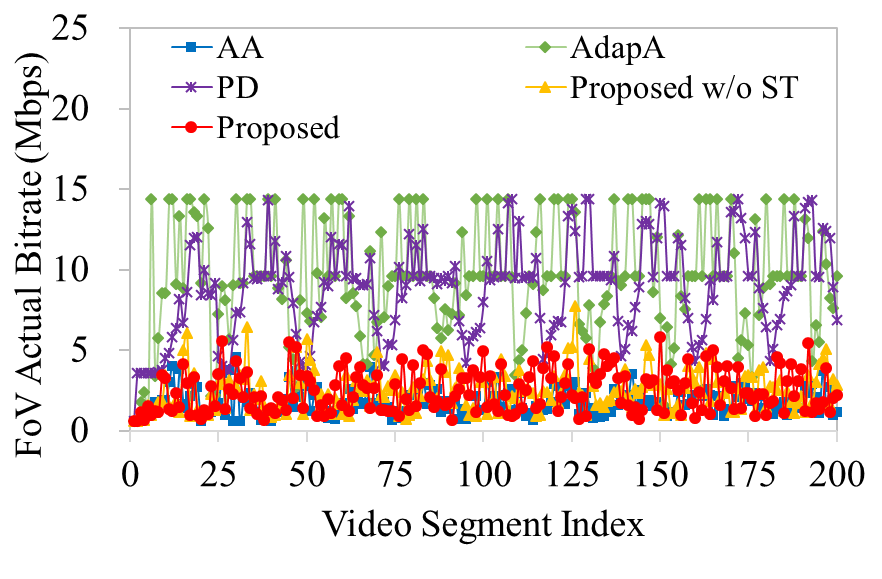}}
\subfigure[]{
\label{fig13:subfig:c}
\includegraphics[width=4cm]{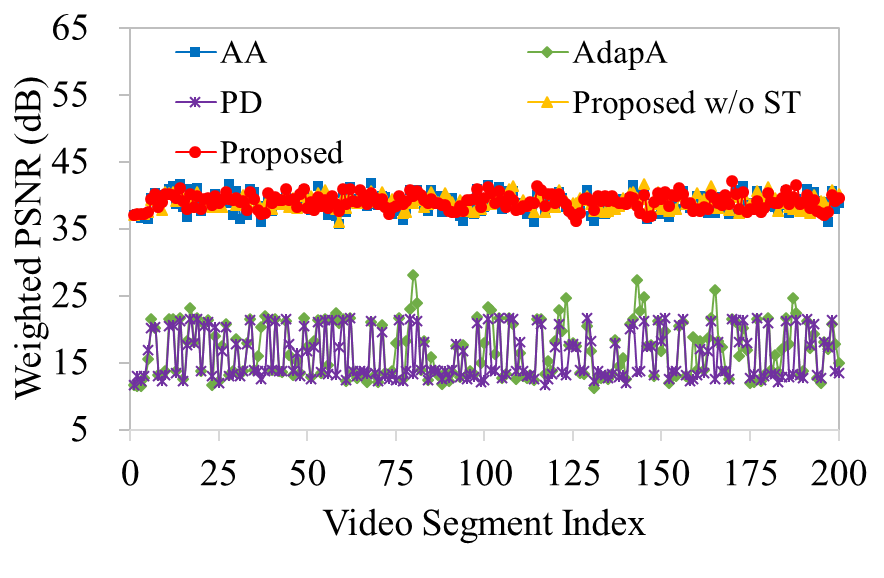}}
\subfigure[]{
\label{fig13:subfig:d}
\includegraphics[width=4cm]{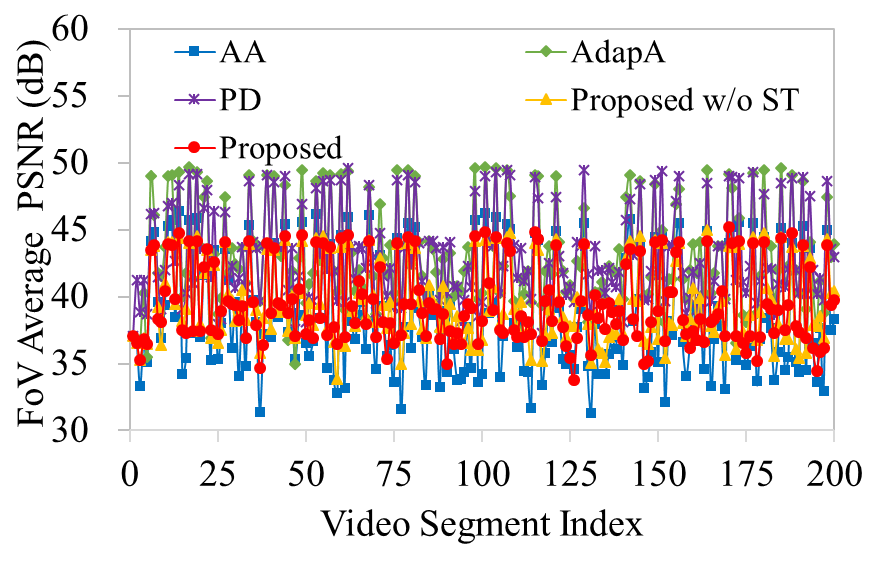}}
\subfigure[]{
\label{fig13:subfig:e}
\includegraphics[width=4cm]{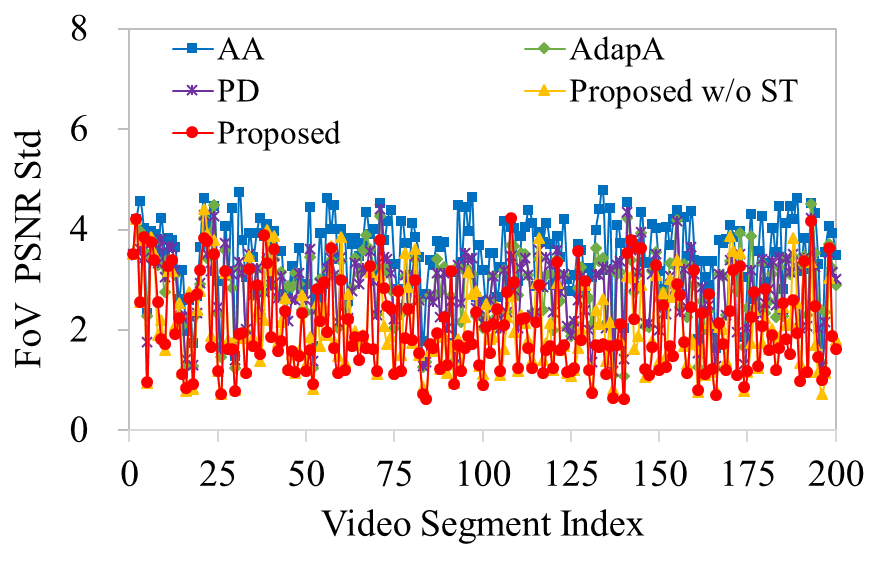}}
\subfigure[]{
\label{fig13:subfig:f}
\includegraphics[width=4cm]{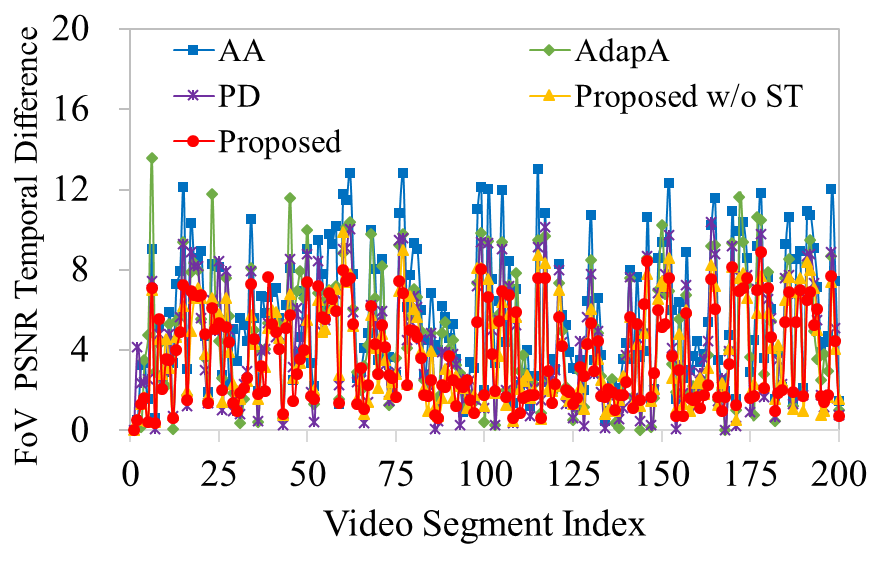}}
\subfigure[]{
\label{fig13:subfig:g}
\includegraphics[width=4cm]{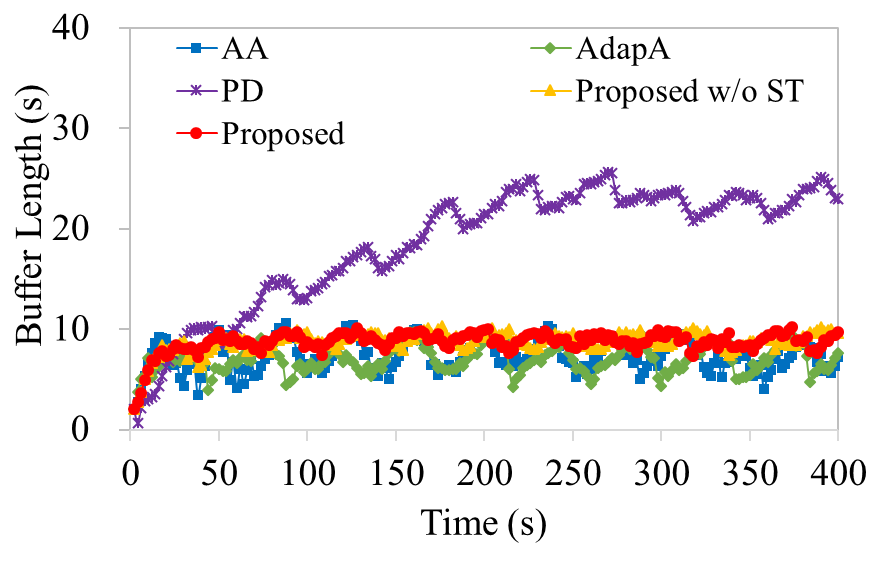}}
\subfigure[]{
\label{fig13:subfig:h}
\includegraphics[width=4cm]{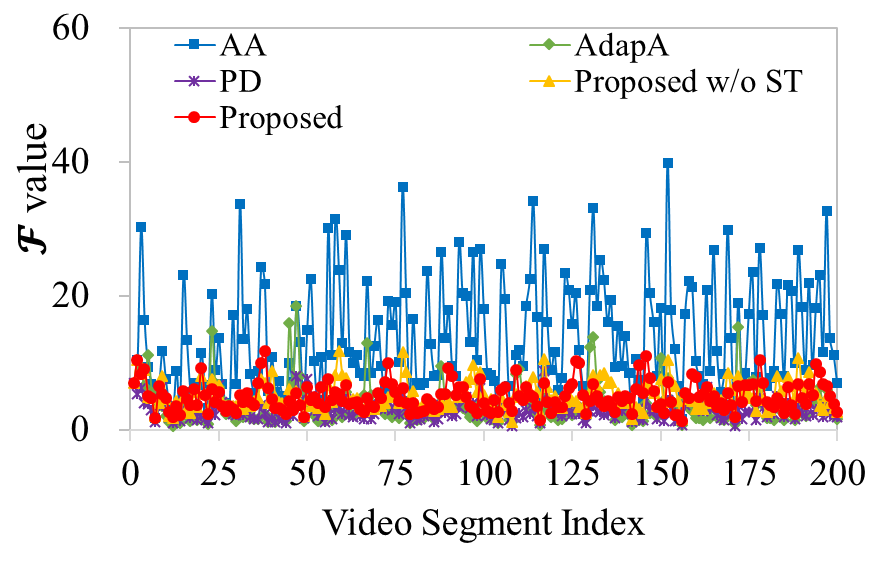}}
\caption{Results of \emph{Case 1} with video sequence \emph{PoleVault}.}
\label{fig13}
\end{figure*}

%表3  QUANTITATIVE COMPARISIONS OF CASE 1 WITH VIDEO SEQUENCE POLEVAULT
% Table generated by Excel2LaTeX from sheet 'PoleVault-虚- 制表-1.2'
\begin{table*}[htbp]
 \scriptsize
  \centering
  \caption{QUANTITATIVE COMPARISIONS OF CASE 1 WITH VIDEO SEQUENCE POLEVAULT}
   \begin{tabular}{cccccccc}
    \toprule
    \toprule
    Sudden view switching  &       & \textit{FoV Actual} & \textit{FoV} & \textit{FoV} & \textit{FoV  PSNR} &       &  \\
    probability in the duration & Method & \textit{Bitrate} & \textit{Average} & \textit{PSNR Std} & \textit{Temporal} & \textit{$\mathcal{F}$ value} & \textit{QoE} \\
    of each video segment &       & \textit{(Mbps)} & \textit{PSNR (dB)} &       & \textit{Difference} &       &  \\
    \midrule
       \multirow{5}[2]{*}{0\%} & AA    & 1.80  & 38.54  & 3.54  & 5.69  & 13.9128  & 2713.72  \\
          & AdapA & \textbf{9.53 } & 43.04  & 2.83  & 4.60  & 3.6132  & 5761.57  \\
          & PD    & 8.81  & \textbf{43.13 } & 2.81  & 4.23  & \textbf{2.8124 } & \textbf{7231.01 } \\
          & Proposed w/o ST & 2.48  & 39.53  & 2.05  & 3.81  & 4.9554  & 6526.02  \\
          & Proposed & 2.45  & 39.50  & \textbf{2.04 } & \textbf{3.69 } & 4.9138  & 6664.61  \\
    \midrule
    \multirow{5}[2]{*}{5\%} & AA    & 1.80  & 38.54  & 3.54  & 5.72  & 13.9212  & 2681.28  \\
          & AdapA & \textbf{9.32 } & 42.13  & 2.93  & 5.43  & 683.0140  & 4584.68  \\
          & PD    & 8.61  & \textbf{42.18 } & 2.91  & 5.11  & 714.0150  & 5991.69  \\
          & Proposed w/o ST & 2.46  & 39.51  & 2.07  & 3.84  & 5.0069  & 6484.86  \\
          & Proposed & 2.43  & 39.48  & \textbf{2.05 } & \textbf{3.72 } & \textbf{4.9570 } & \textbf{6633.89 } \\
    \midrule
    \multirow{5}[2]{*}{10\%} & AA    & 1.81  & 38.61  & 3.53  & 5.71  & 13.8633  & 2715.93  \\
          & AdapA & \textbf{9.11 } & \textbf{41.16 } & 2.98  & 6.30  & 1445.0381  & 3335.20  \\
          & PD    & 8.38  & 41.13  & 2.93  & 6.06  & 1499.5414  & 4644.41  \\
          & Proposed w/o ST & 2.43  & 39.55  & 2.09  & 3.81  & 5.0670  & 6525.18  \\
          & Proposed & 2.40  & 39.52  & \textbf{2.07 } & \textbf{3.72 } & \textbf{5.0551 } & \textbf{6640.70 } \\
    \midrule
    \multirow{5}[2]{*}{20\%} & AA    & 1.81  & 38.54  & 3.51  & 5.65  & 13.7798  & 2767.03  \\
          & AdapA & \textbf{8.71 } & 39.25  & 3.24  & 7.84  & 2882.1543  & 1107.30  \\
          & PD    & 8.05  & 39.24  & 3.18  & 7.62  & 2932.5991  & 2392.20  \\
          & Proposed w/o ST & 2.41  & \textbf{39.45 } & 2.07  & 3.76  & 5.0604  & 6574.42  \\
          & Proposed & 2.39  & 39.42  & \textbf{2.05 } & \textbf{3.64 } & \textbf{4.9961 } & \textbf{6719.47 } \\
    \bottomrule
    \bottomrule
    \end{tabular}%
  \label{tab:addlabel}%
\end{table*}%

%图14
\begin{figure*}
\centering
\subfigure[]{
\label{fig14:subfig:a}
\includegraphics[width=4cm]{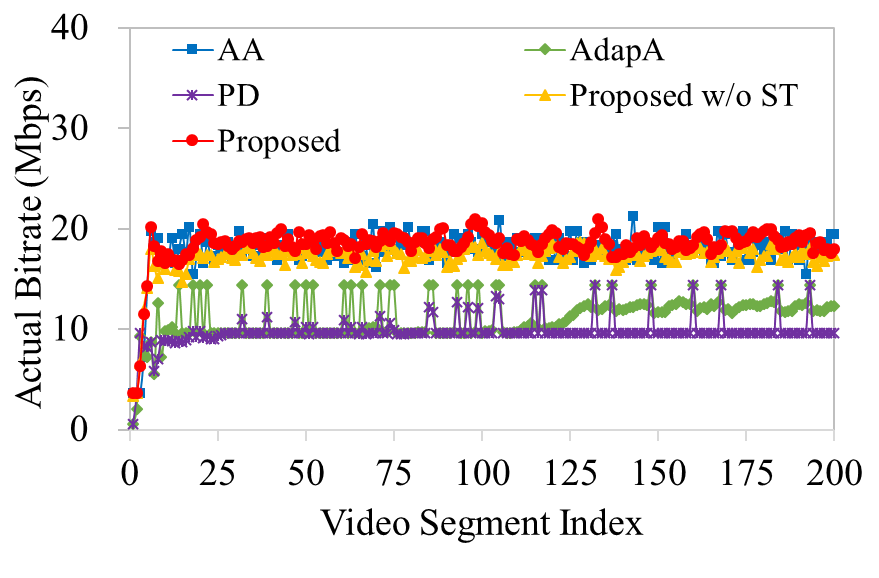}}
\subfigure[]{
\label{fig14:subfig:b}
\includegraphics[width=4cm]{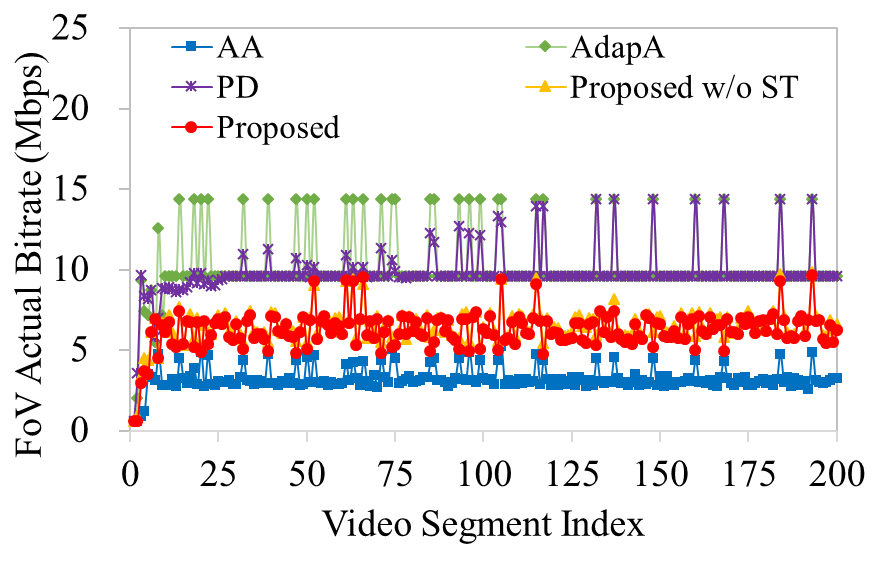}}
\subfigure[]{
\label{fig14:subfig:c}
\includegraphics[width=4cm]{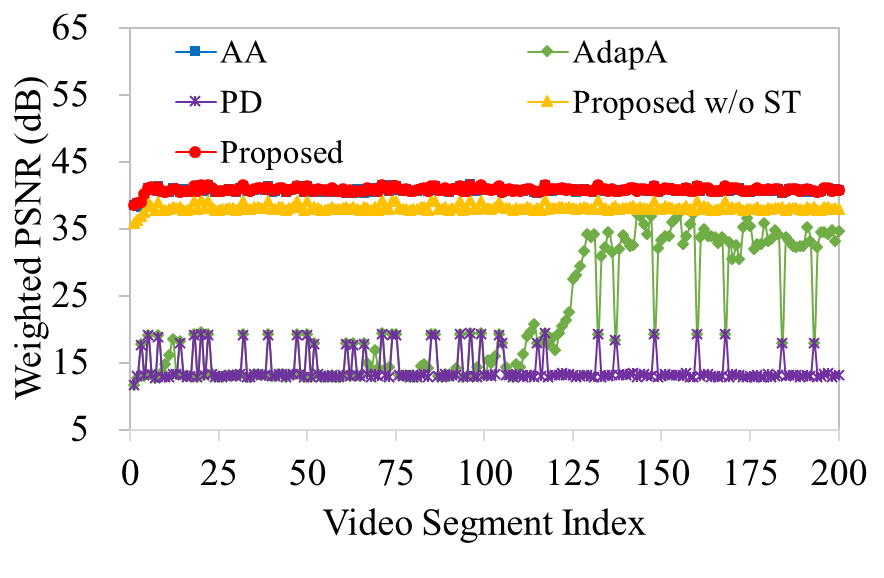}}
\subfigure[]{
\label{fig14:subfig:d}
\includegraphics[width=4cm]{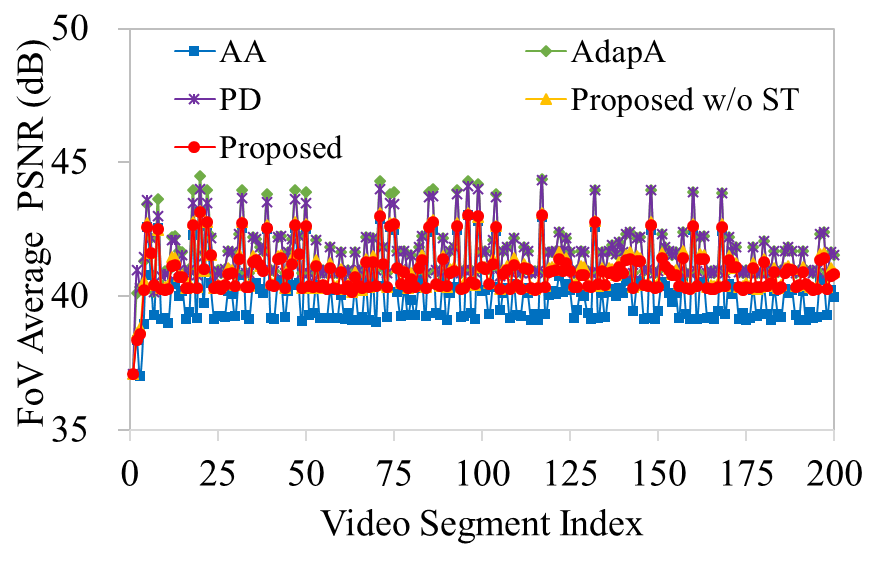}}
\subfigure[]{
\label{fig14:subfig:e}
\includegraphics[width=4cm]{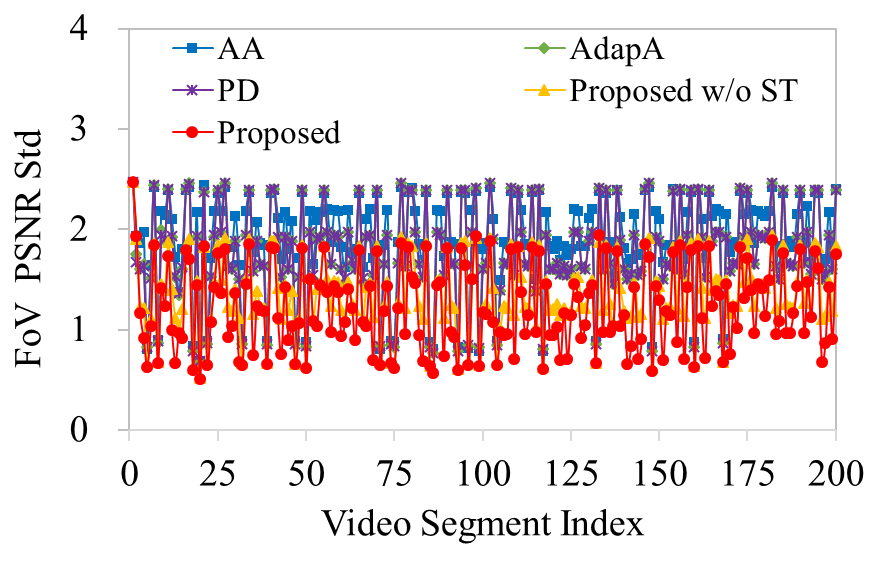}}
\subfigure[]{
\label{fig14:subfig:f}
\includegraphics[width=4cm]{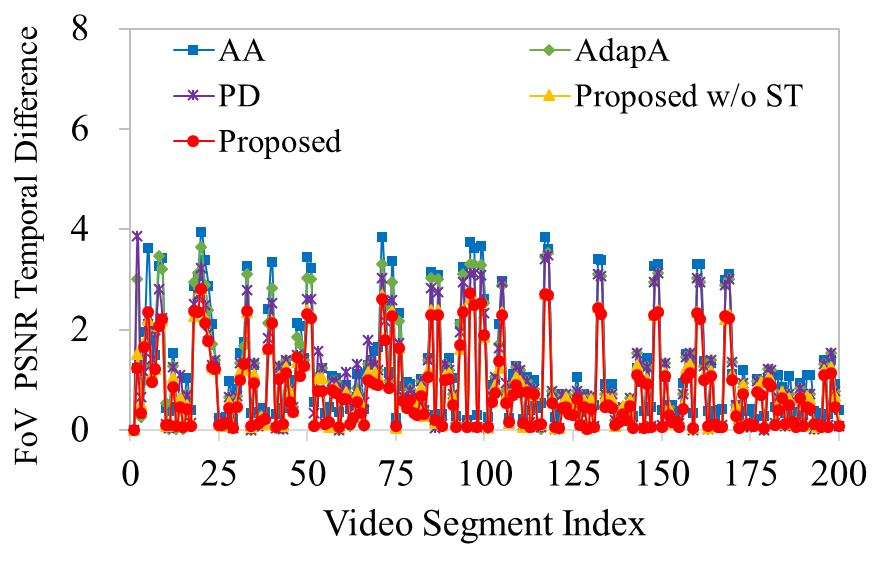}}
\subfigure[]{
\label{fig14:subfig:g}
\includegraphics[width=4cm]{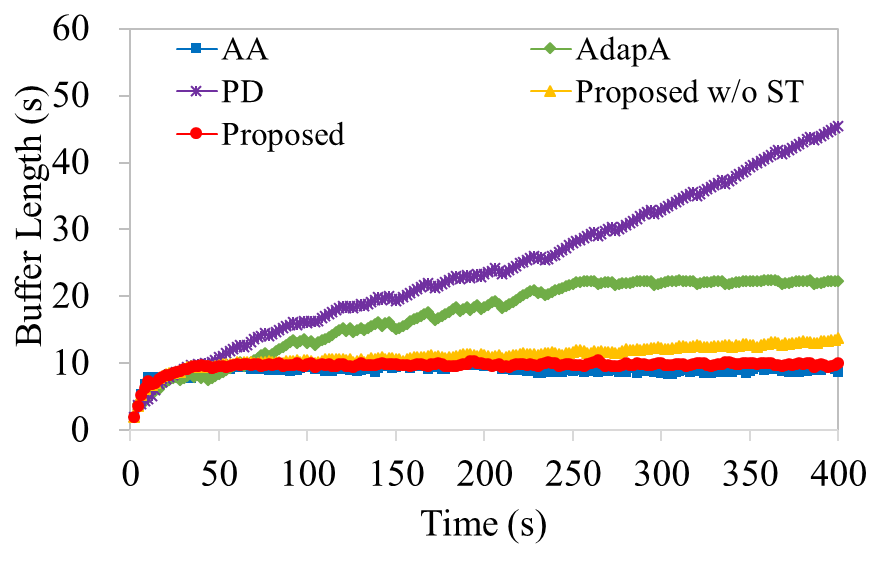}}
\subfigure[]{
\label{fig14:subfig:h}
\includegraphics[width=4cm]{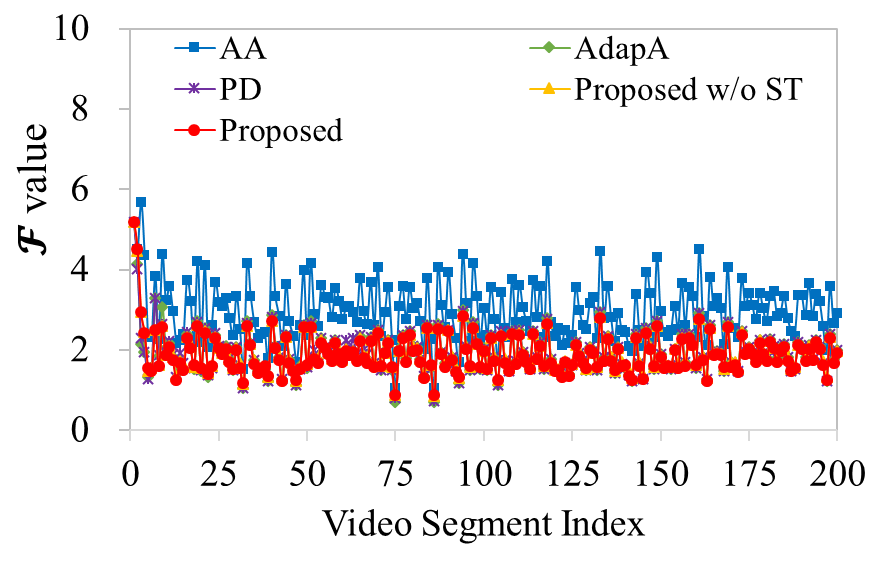}}
\caption{Results of \emph{Case 2} with video sequence \emph{AerialCity}.}
\label{fig14}
\end{figure*}

%表4  QUANTITATIVE COMPARISIONS OF CASE 2 WITH VIDEO SEQUENCE AERIALCITY
% Table generated by Excel2LaTeX from sheet 'AerialCity-台-制表 2.2'
\begin{table*}[htbp]
 \scriptsize
  \centering
  \caption{QUANTITATIVE COMPARISIONS OF CASE 2 WITH VIDEO SEQUENCE AERIALCITY}
    \begin{tabular}{cccccccc}
    \toprule
    \toprule
    Sudden view switching  &       & \textit{FoV Actual} & \textit{FoV} & \textit{FoV} & \textit{FoV  PSNR} &       &  \\
    probability in the duration & Method & \textit{Bitrate} & \textit{Average} & \textit{PSNR Std} & \textit{Temporal} & \textit{$\mathcal{F}$ value} & \textit{QoE} \\
    of each video segment &       & \textit{(Mbps)} & \textit{PSNR (dB)} &       & \textit{Difference} &       &  \\
    \midrule
        \multirow{5}[2]{*}{0\%} & AA    & 3.23  & 40.04  & 1.91  & 1.22  & 2.9870  & 5794.73  \\
          & AdapA & \textbf{10.22 } & \textbf{41.66 } & 1.78  & 1.05  & 1.9365  & 6866.06  \\
          & PD    & 9.80  & 41.62  & 1.78  & 1.03  & 1.9468  & \textbf{6947.33 } \\
          & Proposed w/o ST & 6.13  & 40.91  & 1.31  & 0.84  & 1.9346  & 6592.22  \\
          & Proposed & 6.24  & 40.89  & \textbf{1.22 } & \textbf{0.79 } & \textbf{1.8986 } & 6643.07  \\
    \midrule
    \multirow{5}[2]{*}{5\%} & AA    & 3.26  & 40.09  & 1.89  & 1.24  & 2.9655  & 5781.74  \\
          & AdapA & \textbf{10.02 } & \textbf{41.08 } & 1.90  & 1.61  & 477.2496  & 6076.49  \\
          & PD    & 9.59  & 40.72  & 1.93  & 1.91  & 730.0694  & 5712.55  \\
          & Proposed w/o ST & 6.06  & 40.92  & 1.30  & 0.85  & 1.9307  & 6581.81  \\
          & Proposed & 6.16  & 40.90  & \textbf{1.22 } & \textbf{0.81 } & \textbf{1.8960 } & \textbf{6629.96 } \\
    \midrule
    \multirow{5}[2]{*}{10\%} & AA    & 3.28  & 40.11  & 1.88  & 1.26  & 2.9546  & 5768.19  \\
          & AdapA & \textbf{9.79 } & 40.33  & 2.05  & 2.36  & 1094.3622  & 5035.36  \\
          & PD    & 9.35  & 39.63  & 1.88  & 2.99  & 1506.4761  & 4194.56  \\
          & Proposed w/o ST & 5.99  & \textbf{40.91 } & 1.29  & 0.86  & 1.9352  & 6568.95  \\
          & Proposed & 6.10  & 40.90  & \textbf{1.21 } & \textbf{0.82 } & \textbf{1.9005 } & \textbf{6617.84 } \\
    \midrule
    \multirow{5}[2]{*}{20\%} & AA    & 3.30  & 40.12  & 1.86  & 1.23  & 2.9321  & 5797.00  \\
          & AdapA & \textbf{9.31 } & 38.69  & 2.22  & 3.90  & 2350.6802  & 2856.96  \\
          & PD    & 8.90  & 37.78  & 1.98  & 4.73  & 2888.0340  & 1737.18  \\
          & Proposed w/o ST & 5.86  & \textbf{40.87 } & 1.28  & 0.85  & 1.9449  & 6568.43  \\
          & Proposed & 5.97  & 40.86  & \textbf{1.20 } & \textbf{0.81 } & \textbf{1.9096 } & \textbf{6616.21 } \\
    \bottomrule
    \bottomrule
    \end{tabular}%
  \label{tab:addlabel}%
\end{table*}%

%图15
\begin{figure*}
\centering
\subfigure[]{
\label{fig15:subfig:a}
\includegraphics[width=4cm]{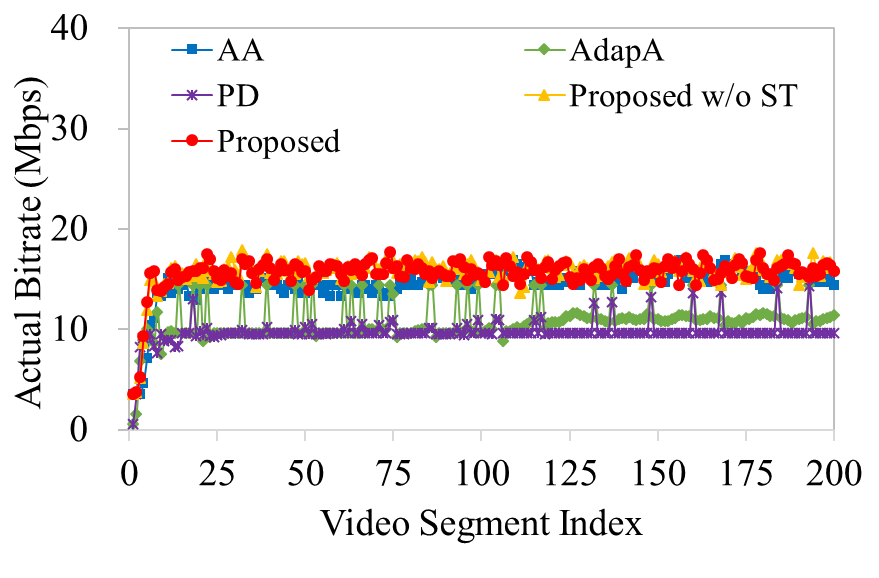}}
\subfigure[]{
\label{fig15:subfig:b}
\includegraphics[width=4cm]{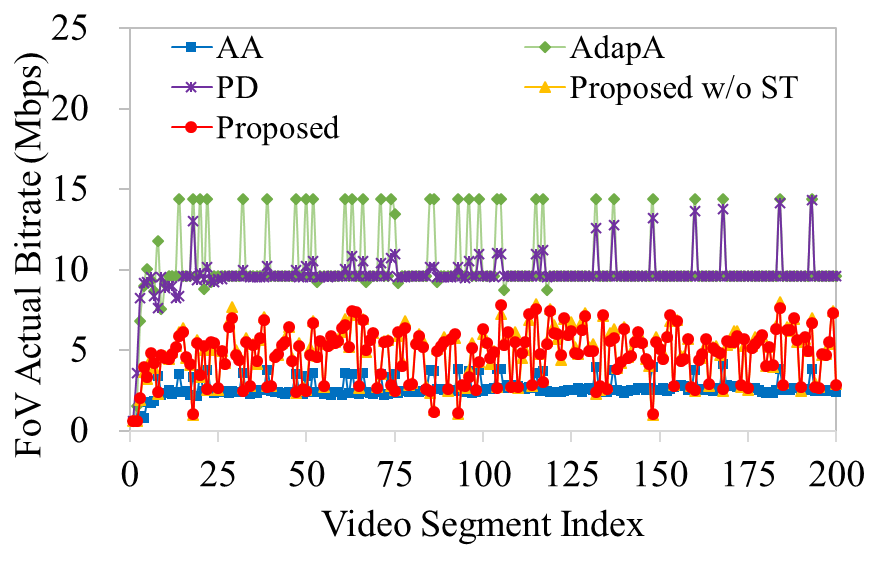}}
\subfigure[]{
\label{fig15:subfig:c}
\includegraphics[width=4cm]{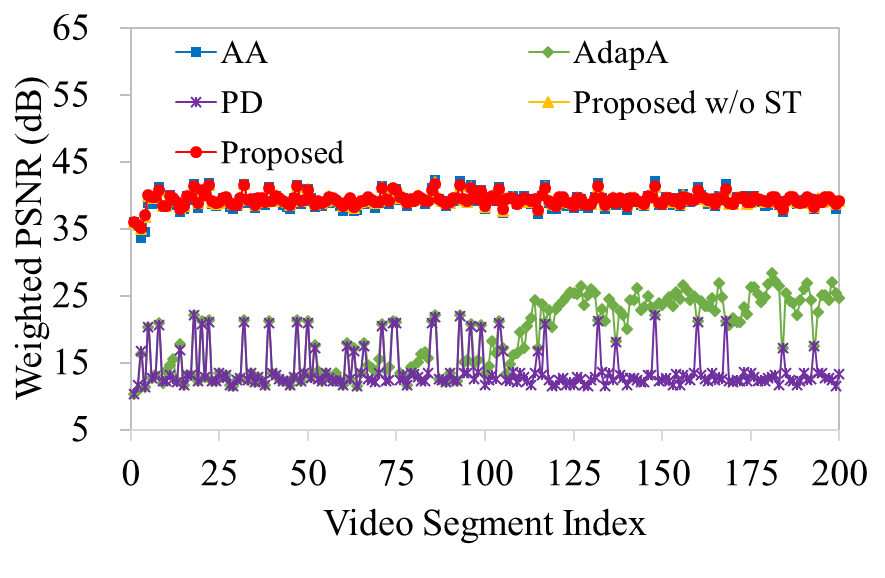}}
\subfigure[]{
\label{fig15:subfig:d}
\includegraphics[width=4cm]{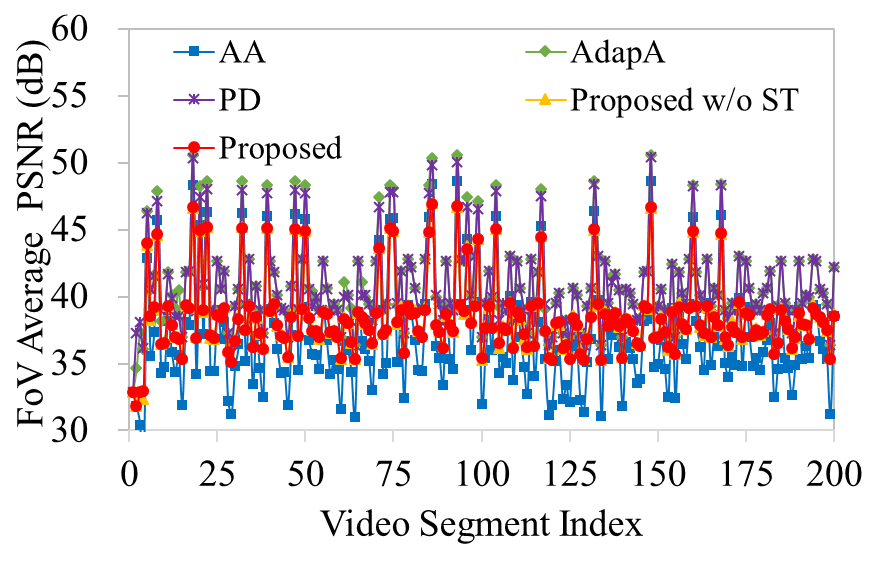}}
\subfigure[]{
\label{fig15:subfig:e}
\includegraphics[width=4cm]{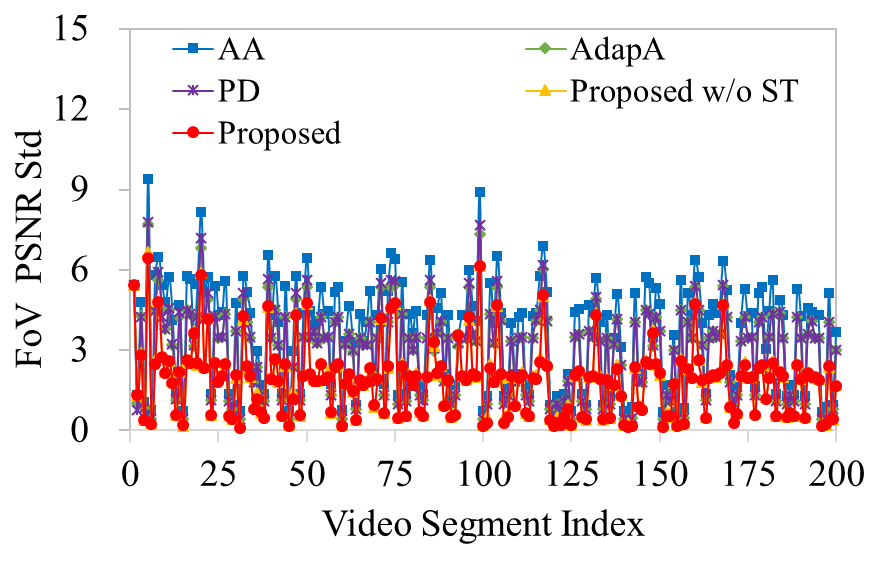}}
\subfigure[]{
\label{fig15:subfig:f}
\includegraphics[width=4cm]{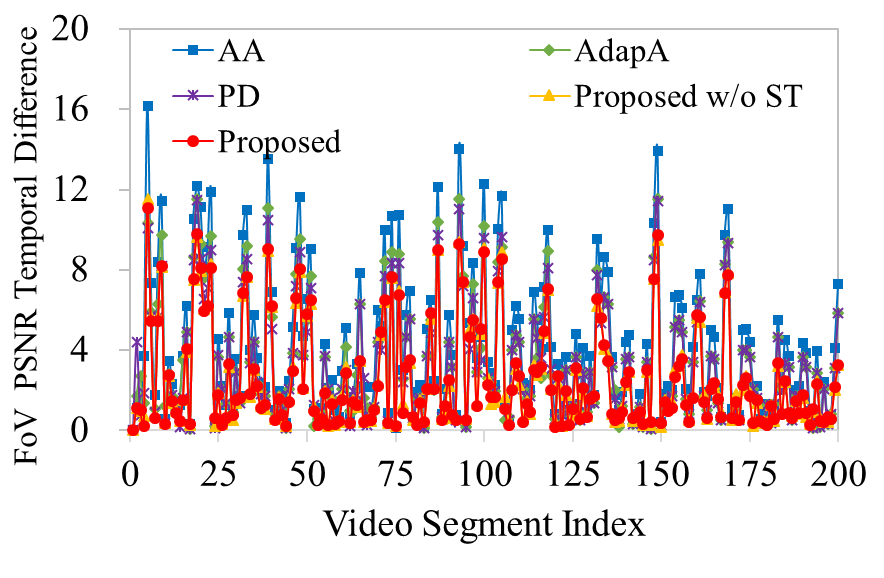}}
\subfigure[]{
\label{fig15:subfig:g}
\includegraphics[width=4cm]{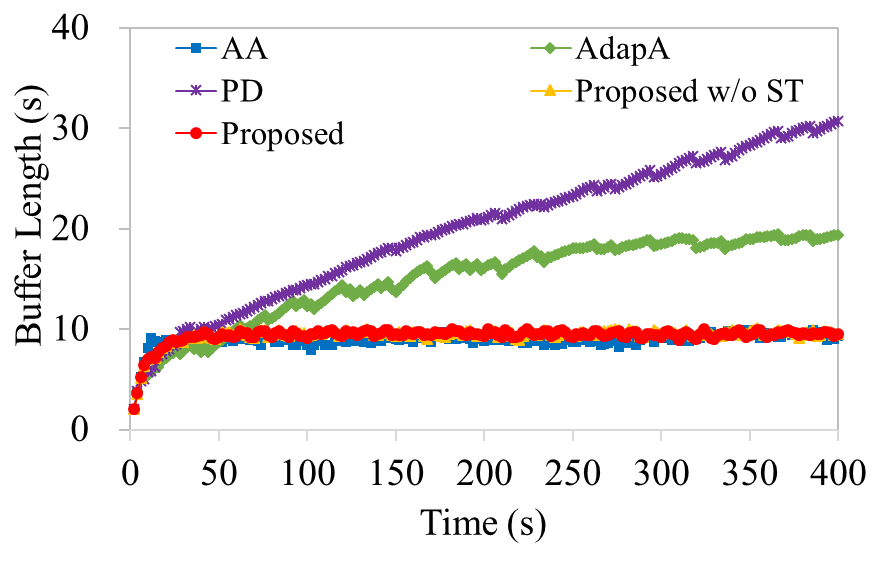}}
\subfigure[]{
\label{fig15:subfig:h}
\includegraphics[width=4cm]{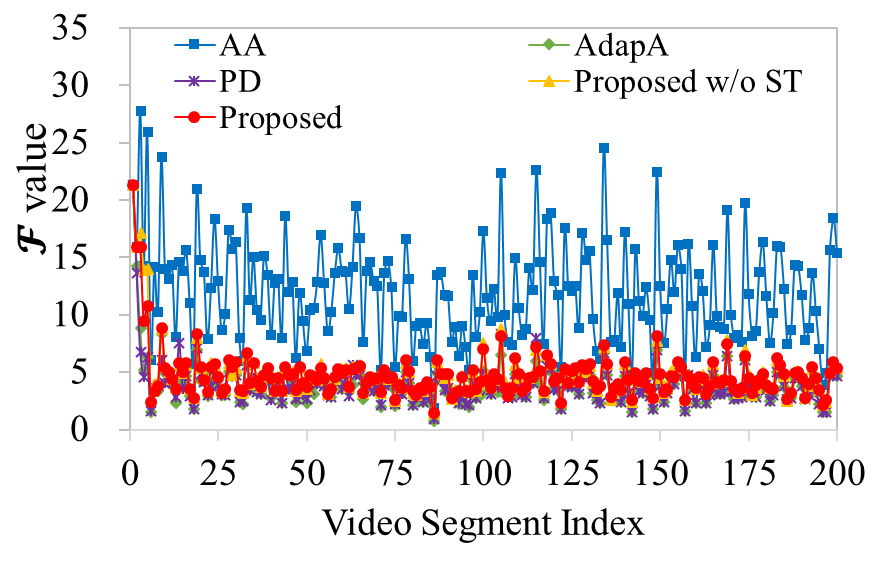}}
\caption{Results of \emph{Case 2} with video sequence \emph{DrivingInCountry}.}
\label{fig15}
\end{figure*}

%表5  QUANTITATIVE COMPARISIONS OF CASE 2 WITH VIDEO SEQUENCE DRIVINGINCOUNTRY
% Table generated by Excel2LaTeX from sheet 'DrivingInCountry- 台-制表 2.2'
\begin{table*}[htbp]
 \scriptsize
  \centering
  \caption{QUANTITATIVE COMPARISIONS OF CASE 2 WITH VIDEO SEQUENCE DRIVINGINCOUNTRY}
    \begin{tabular}{cccccccc}
    \toprule
    \toprule
    Sudden view switching  &       & \textit{FoV Actual} & \textit{FoV} & \textit{FoV} & \textit{FoV  PSNR} &       &  \\
    probability in the duration & Method & \textit{Bitrate} & \textit{Average} & \textit{PSNR Std} & \textit{Temporal} & \textit{$\mathcal{F}$ value} & \textit{QoE} \\
    of each video segment &       & \textit{(Mbps)} & \textit{PSNR (dB)} &       & \textit{Difference} &       &  \\
    \midrule
        \multirow{5}[2]{*}{0\%} & AA    & 2.63  & 36.87  & 3.67  & 4.18  & 12.4804  & 6276.68  \\
          & AdapA & \textbf{10.21 } & \textbf{40.87 } & 2.98  & 3.37  & \textbf{3.8464 } & 7757.06  \\
          & PD    & 9.71  & 40.79  & 3.00  & 3.27  & 3.8911  & \textbf{8101.26 } \\
          & Proposed w/o ST & 4.75  & 38.45  & 1.89  & 2.59  & 4.7437  & 7948.48  \\
          & Proposed & 4.70  & 38.41  & \textbf{1.86 } & \textbf{2.58 } & 4.7083  & 7952.63  \\
    \midrule
    \multirow{5}[2]{*}{5\%} & AA    & 2.66  & 37.06  & 3.72  & 4.25  & 12.4060  & 6228.81  \\
          & AdapA & \textbf{10.01 } & \textbf{40.24 } & 3.11  & 3.99  & 570.8212  & 6886.79  \\
          & PD    & 9.51  & 39.95  & 3.12  & 4.06  & 731.9261  & 6979.09  \\
          & Proposed w/o ST & 4.68  & 38.56  & 1.96  & 2.64  & 4.7545  & 7904.02  \\
          & Proposed & 4.63  & 38.52  & \textbf{1.93 } & \textbf{2.63 } & \textbf{4.7235 } & \textbf{7909.38 } \\
    \midrule
    \multirow{5}[2]{*}{10\%} & AA    & 2.68  & 37.17  & 3.75  & 4.31  & 12.4351  & 6177.52  \\
          & AdapA & \textbf{9.77 } & \textbf{39.41 } & 3.21  & 4.81  & 1266.0067  & 5737.11  \\
          & PD    & 9.27  & 38.91  & 3.08  & 5.09  & 1508.2876  & 5533.58  \\
          & Proposed w/o ST & 4.63  & 38.61  & 1.99  & 2.71  & 4.8424  & 7833.55  \\
          & Proposed & 4.58  & 38.57  & \textbf{1.96 } & \textbf{2.70 } & \textbf{4.8018 } & \textbf{7840.25 } \\
    \midrule
    \multirow{5}[2]{*}{20\%} & AA    & 2.69  & 36.99  & 3.71  & 4.18  & 12.5354  & 6290.44  \\
          & AdapA & \textbf{9.30 } & 37.54  & 3.26  & 6.24  & 2608.7359  & 3652.25  \\
          & PD    & 8.83  & 36.99  & 3.07  & 6.58  & 2889.6746  & 3369.29  \\
          & Proposed w/o ST & 4.65  & \textbf{38.44 } & 1.98  & 2.66  & 4.9926  & 7864.77  \\
          & Proposed & 4.61  & 38.40  & \textbf{1.96 } & \textbf{2.65 } & \textbf{4.9554 } & \textbf{7864.82 } \\
    \bottomrule
    \bottomrule
    \end{tabular}%
  \label{tab:addlabel}%
\end{table*}%

%图16
\begin{figure*}
\centering
\subfigure[]{
\label{fig16:subfig:a}
\includegraphics[width=4cm]{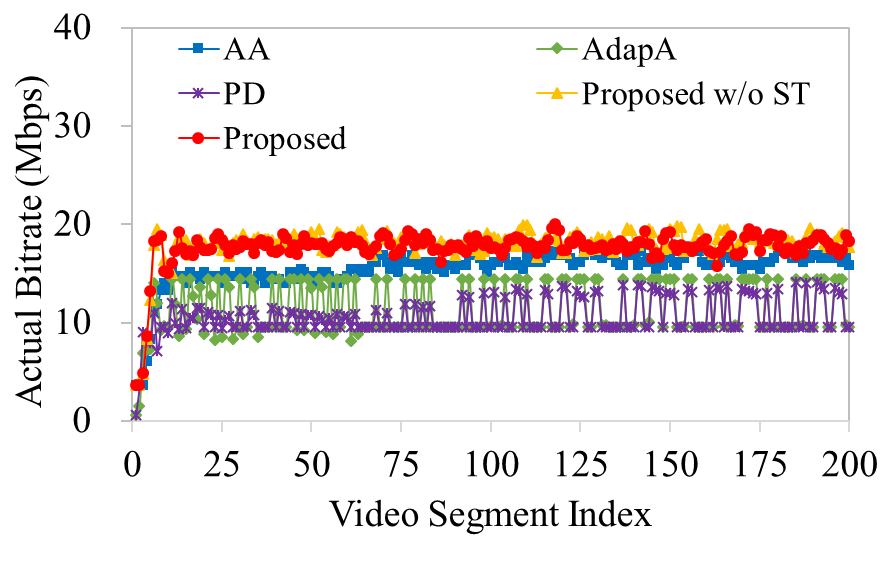}}
\subfigure[]{
\label{fig16:subfig:b}
\includegraphics[width=4cm]{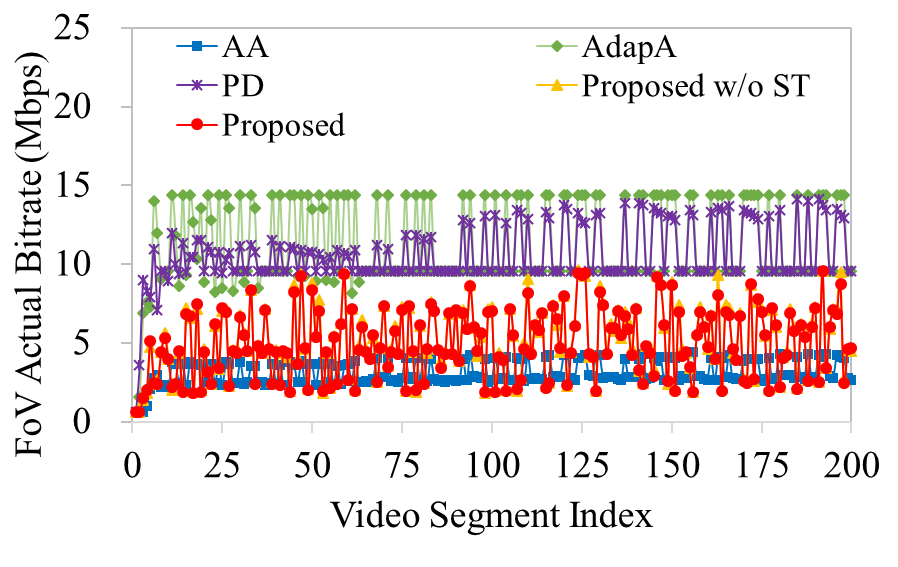}}
\subfigure[]{
\label{fig16:subfig:c}
\includegraphics[width=4cm]{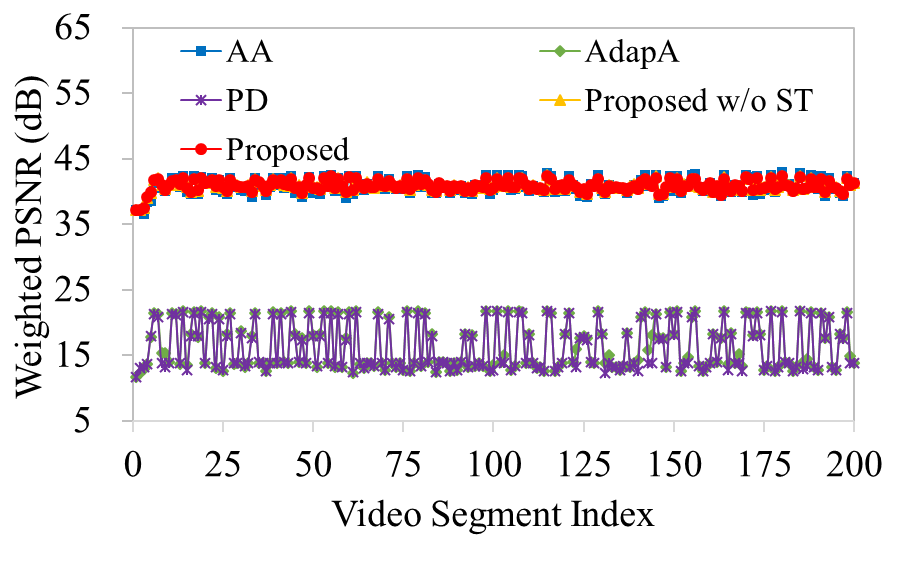}}
\subfigure[]{
\label{fig16:subfig:d}
\includegraphics[width=4cm]{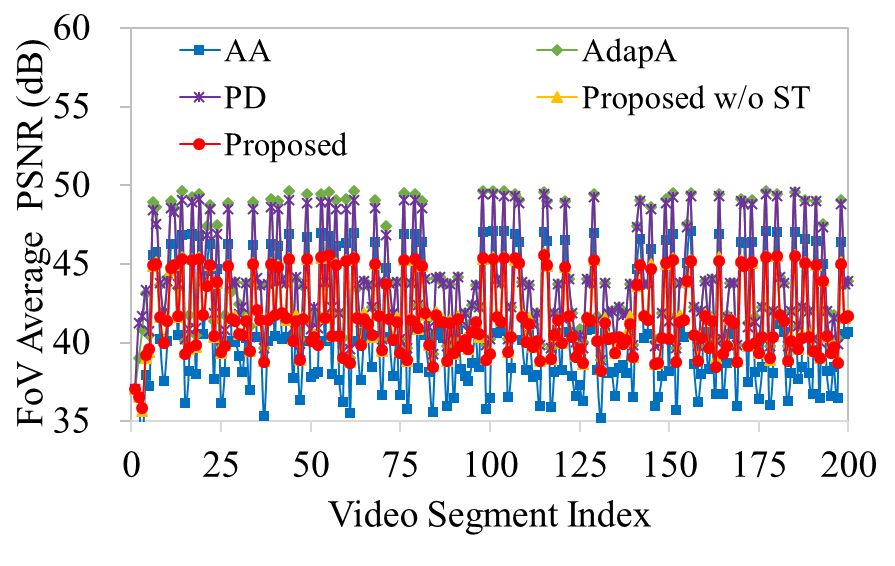}}
\subfigure[]{
\label{fig16:subfig:e}
\includegraphics[width=4cm]{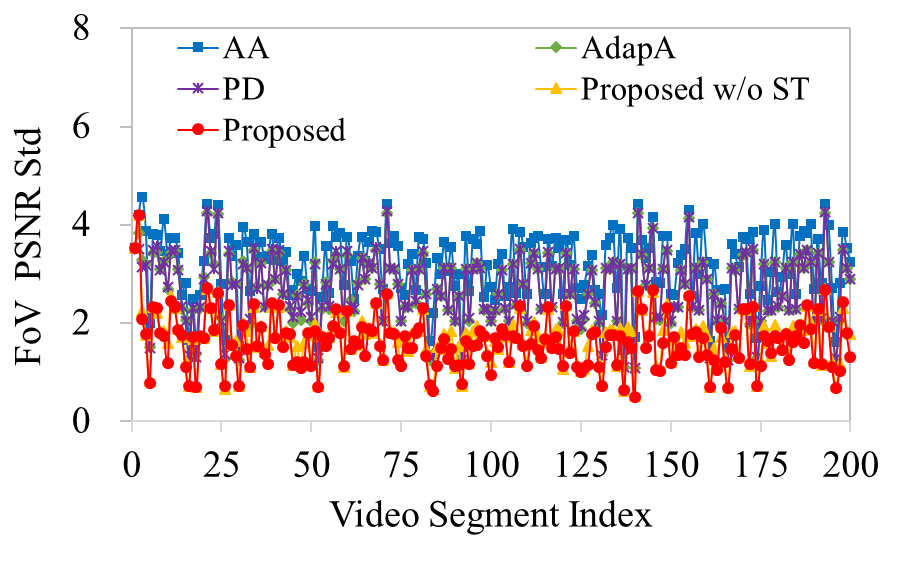}}
\subfigure[]{
\label{fig16:subfig:f}
\includegraphics[width=4cm]{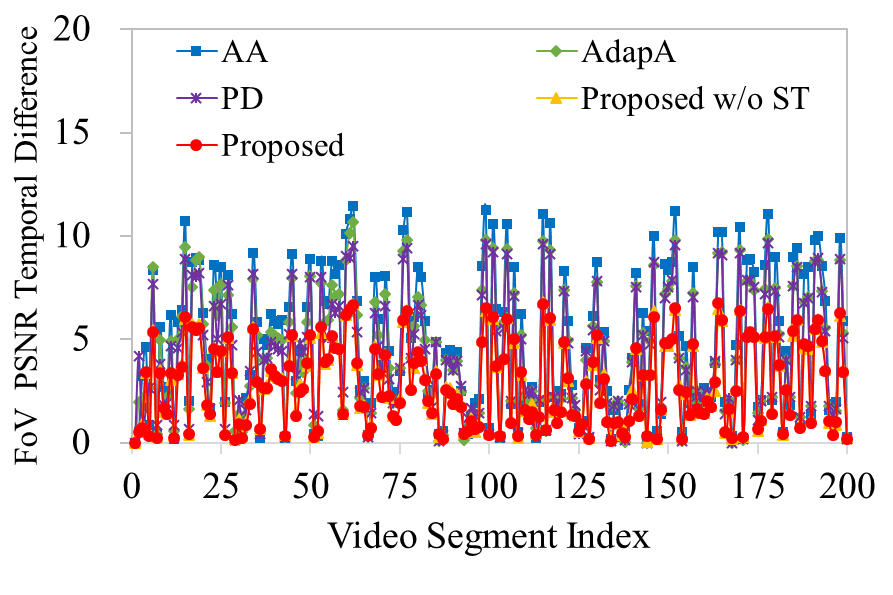}}
\subfigure[]{
\label{fig16:subfig:g}
\includegraphics[width=4cm]{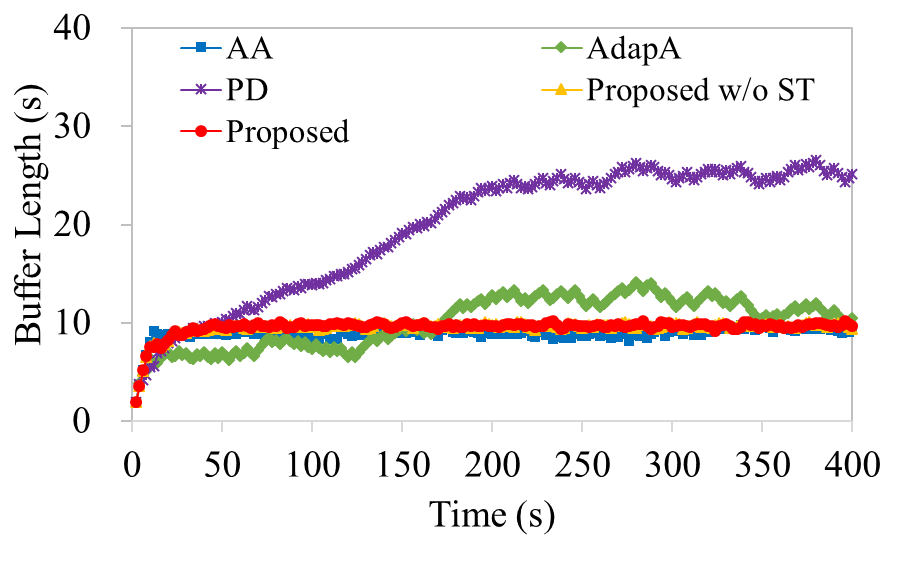}}
\subfigure[]{
\label{fig16:subfig:h}
\includegraphics[width=4cm]{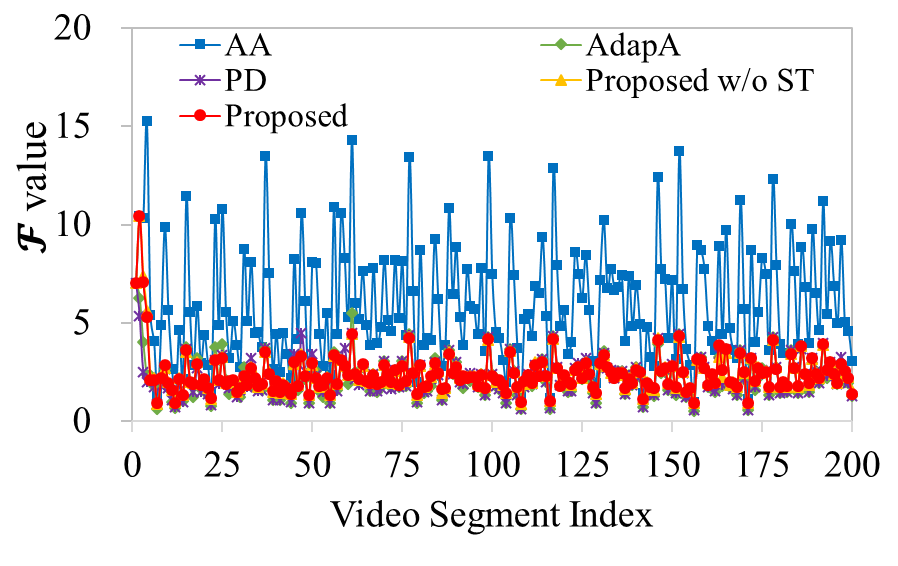}}
\caption{Results of \emph{Case 2} with video sequence \emph{PoleVault}.}
\label{fig16}
\end{figure*}

%表6   QUANTITATIVE COMPARISIONS OF CASE 2 WITH VIDEO SEQUENCE POLEVAULT
% Table generated by Excel2LaTeX from sheet 'PoleVault-台- 制表 2.2'
\begin{table*}[htbp]
 \scriptsize
  \centering
  \caption{QUANTITATIVE COMPARISIONS OF CASE 2 WITH VIDEO SEQUENCE POLEVAULT}
   \begin{tabular}{cccccccc}
    \toprule
    \toprule
    Sudden view switching  &       & \textit{FoV Actual} & \textit{FoV} & \textit{FoV} & \textit{FoV  PSNR} &       &  \\
    probability in the duration & Method & \textit{Bitrate} & \textit{Average} & \textit{PSNR Std} & \textit{Temporal} & \textit{$\mathcal{F}$ value} & \textit{QoE} \\
    of each video segment &       & \textit{(Mbps)} & \textit{PSNR (dB)} &       & \textit{Difference} &       &  \\
    \midrule
        \multirow{5}[2]{*}{0\%} & AA    & 3.15  & 40.54  & 3.21  & 4.75  & 6.1603  & 6331.66  \\
          & AdapA & \textbf{11.35 } & \textbf{43.88 } & 2.70  & 4.22  & \textbf{2.2275 } & 7080.98  \\
          & PD    & 10.61  & 43.77  & 2.71  & 4.12  & 2.2429  & 7663.80  \\
          & Proposed w/o ST & 5.04  & 41.59  & 1.68  & 2.78  & 2.4270  & 8375.75  \\
          & Proposed & 4.95  & 41.51  & \textbf{1.61 } & \textbf{2.76 } & 2.4235  & \textbf{8399.94 } \\
    \midrule
    \multirow{5}[2]{*}{5\%} & AA    & 3.16  & 40.54  & 3.21  & 4.78  & 6.1704  & 6298.79  \\
          & AdapA & \textbf{11.09 } & \textbf{42.91 } & 2.80  & 5.11  & 713.4498  & 5823.29  \\
          & PD    & 10.37  & 42.81  & 2.82  & 5.00  & 713.4641  & 6408.60  \\
          & Proposed w/o ST & 5.00  & 41.56  & 1.69  & 2.80  & 2.4512  & 8346.05  \\
          & Proposed & 4.92  & 41.48  & \textbf{1.62 } & \textbf{2.78 } & \textbf{2.4492 } & \textbf{8368.39 } \\
    \midrule
    \multirow{5}[2]{*}{10\%} & AA    & 3.17  & 40.61  & 3.21  & 4.75  & 6.1574  & 6344.58  \\
          & AdapA & \textbf{10.82 } & \textbf{41.87 } & 2.84  & 6.07  & 1496.2501  & 4466.65  \\
          & PD    & 10.11  & 41.75  & 2.84  & 5.97  & 1499.0027  & 5032.74  \\
          & Proposed w/o ST & 4.92  & 41.57  & 1.70  & 2.78  & 2.4608  & 8373.40  \\
          & Proposed & 4.84  & 41.49  & \textbf{1.64 } & \textbf{2.76 } & \textbf{2.4589 } & \textbf{8393.66 } \\
    \midrule
    \multirow{5}[2]{*}{20\%} & AA    & 3.17  & 40.54  & 3.19  & 4.70  & 6.1193  & 6394.97  \\
          & AdapA & \textbf{10.34 } & 39.94  & 3.10  & 7.65  & 2917.8390  & 2178.77  \\
          & PD    & 9.68  & 39.82  & 3.11  & 7.57  & 2932.1145  & 2727.61  \\
          & Proposed w/o ST & 4.89  & \textbf{41.47 } & 1.69  & 2.76  & 2.4871  & 8382.43  \\
          & Proposed & 4.81  & 41.39  & \textbf{1.62 } & \textbf{2.74 } & \textbf{2.4867 } & \textbf{8402.37 } \\
    \bottomrule
    \bottomrule
    \end{tabular}%
  \label{tab:addlabel}%
\end{table*}%

When comparing the \emph{FoV PSNR Std} shown in Fig. 11(e), we can see that the \emph{FoV PSNR Std} of the \textbf{Proposed Method} is the smallest. For the \emph{FoV PSNR Temporal Difference}, the \textbf{PD method} is the smallest (followed by the \textbf{Proposed Method}) for \emph{AerialCity} and \emph{DrivingInCountry}, while that of the \textbf{Proposed Method} is the smallest for \emph{PoleVault}. The reason is that the \textbf{PD method} always guarantees the highest FoV quality of all the video segments by ignoring the other tiles. Moreover, from Fig. 11(g), the \emph{buffer length} of the \textbf{Proposed Method} is more stable than the other methods, and the buffer of the \textbf{PD Method} continues to grow because it uses the bandwidth inadequately.

When comparing the $\mathcal{F}$ \emph{value} shown in Fig. 11(h), the \textbf{PD Method} is the smallest, while that of the \textbf{AA Method} is the largest. Note that Fig.11 is obtained under the experiments that there is no sudden view switching in the display duration of each video segment. Therefore, the $\mathcal{F}$ \emph{value} is unfair for the proposed method. Consider the circumstance that a user is watching a video segment at the current viewport. The tiles of the current video segment are downloaded based on the current viewport. During the playback of the current video segment, the user may switch his/her viewing angle suddenly even if the viewport has been accurately predicted at the beginning of the current video segment. To simulate this sudden change, we randomly select $n$ video segments (from a total $N$ video segments) during which a viewport switching occurs. The sudden view switching probability is calculated to be  $p=\frac{n}{N}$. In the experiment, we take the value of $p$ as 5\%, 10\% and 20\% and we can observe that the \emph{FoV PSNR Std}, \emph{FoV PSNR Temporal Difference} and $\mathcal{F}$ \emph{value} of the \textbf{Proposed Method} are always the smallest. Since the user's FoV may be switched to other tiles that were not downloaded in the \textbf{PD Method}, its $\mathcal{F}$ \emph{value} is the largest.

Besides, we also evaluated the performance of different methods by using a commonly used \emph{QoE} metric [27]:
\begin{equation}\label{E16}
   \begin{array}{l}
    QoE = \sum\nolimits_{l = 1}^L {{q_l}}  - \gamma \sum\nolimits_{l = 1}^{L - 1} {\left| {{q_{l + 1}} - {q_l}} \right|} \vspace{1ex}\\
    \quad \qquad - \delta \sum\nolimits_{l = 1}^L {\max \left[ {0,{t_{download,l}} - {b_l}} \right]} \vspace{1ex}\\
    \quad \qquad - \eta {\sum\nolimits_{l = 1}^{L - 1} {\left( {\max \left[ {0,{b_{ref}} - {b_{l + 1}}} \right]} \right)} ^2},
\end{array}
\end{equation}
where $\gamma$=6, $\delta$=500 and $\eta$=0.1 are model parameters that are empirically defined [27]. $L$ is the number of received segments, ${q_l}$ is the \emph{FoV average PSNR} value of $l$-th segment, ${t_{download,l}}$ is the download time of $l$-th segment, ${b_l}$ is the buffer length at the end time of the $l$-th segment, and ${b_{ref}}$=15 seconds. Note that ${q_l}$ is only calculated by the average PSNR of tiles in the FoV. The detailed quantitative comparisons are also provided in Table I, from which we can see that the user \emph{QoE} of the \textbf{Proposed Method} is the best.

Similar results can also be found in Figs.12 and 13 and Tables II and III, for \emph{DrivingInCounrty} and \emph{PoleVault}. This means that the \textbf{Proposed Method} is robust to various video content.\vspace{1ex}

\noindent \emph{(ii) Results of Case 2} \vspace{1ex}%IV--B--2

In the case of a real network environment, Figs. 14-15 compare the performance of the bit allocation methods. Taking \emph{AerialCity} as an example, from Fig. 14(a), we can observe that the \emph{Actual Bitrates} of the \textbf{AdpaA Method} and the \textbf{PD Method} are smaller than the other methods.

Similar to the results of \emph{Case 1}, Fig. 14(b) shows that the \emph{FoV Actual Bitrates} of the \textbf{AdpaA Method} and the \textbf{PD Method} are the larger because it distributes more bitrate to the current FoV, while that of the \textbf{AA Method} is smaller. In Fig. 14 (c), the \emph{Weighted PSNR} of the \textbf{Proposed Method} is the highest, while this value of the \textbf{PD Method} is the smallest. Although, as shown in Fig. 14(d), the \emph{FoV Average PSNR} of the \textbf{Proposed Method} is not the highest, the spatial and temporal smoothness are the best as shown in Figs. 14(e) and (f). From Table IV, we can see that when there is no sudden view switching during the display of each video segments, the \emph{QoE} of the \textbf{PD Method} is the largest, and the $\mathcal{F}$ \emph{value} of it still very small; whereas, in this case, the \emph{QoE} of the \textbf{Proposed Method} is also large, and the $\mathcal{F}$ \emph{value} of the \textbf{Proposed Method} is the smallest. When there exists sudden view switching during the display of some video segments, the $\mathcal{F}$ \emph{values} of the \textbf{Proposed Method} are still the smallest, meanwhile the corresponding \emph{QoEs} are also the largest. Similar results can also be found for the other two tested video sequences, \emph{DrivingInCountry} and \emph{PoleVault}.

\section{Conclusion} % V
In this paper, we have presented an effective adaptive streaming framework for 360 videos. We first presented a novel bitrate adaptation algorithm for 360-degree videos to determine the requested bitrate. Then, the Gaussian model is adopted to predict the FoV at the starting time of each requested video segment. Besides, to tackle the risk that the view angle is switched during the display of a video segment, all tiles in the 360-degree video are downloaded. Because users can only watch the content of the FoV in a 360-degree video, a Zipf model is proposed to determine priorities for different tiles. Finally, a two-stage coarse to fine optimization algorithm is proposed to allocate bitrates for all the tiles so that the video quality as well as the spatial and temporal smoothness of tiles in the FoV can be preserved. Experimental results show that the performance of our proposed method is much better than the state-of-the-art methods.

\section*{Acknowledgment}
The authors would like to thank Institute of Information Technology (ITEC) at Klagenfurt University for the valuable and basis work of DASH. They would also like to thank the editors and anonymous reviewers for their valuable comments.

\ifCLASSOPTIONcaptionsoff
  \newpage
\fi


%\end{IEEEbiography}

\begin{IEEEbiography}
[{\includegraphics[width=1in,height=1.2in,clip,keepaspectratio]{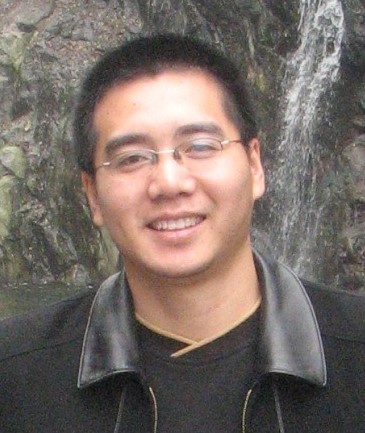}}]{Hui Yuan} (S'08-M'12-SM'17) received the B.E. and Ph.D. degree in telecommunication engineering from Xidian University, Xi'an, China, in 2006 and 2011, respectively. From 2011.04 to now, he works as a lecturer (2011.04-2014.12), an associate Professor (2015.01-2016.08), and a full professor (2016.09-), at Shandong University (SDU), Jinan, China. From 2013.01-2014.12, and 2017.11-2018.02, he also worked as a post-doctor fellow (Granted by the Hong Kong Scholar Project) and a research fellow, respectively, with the department of computer science, City University of Hong Kong (CityU). His current research interests include video/image/immersive media processing, compression, adaptive streaming, and computer vision, etc.
\end{IEEEbiography}

\begin{IEEEbiography}
[{\includegraphics[width=1in,height=1.2in,clip,keepaspectratio]{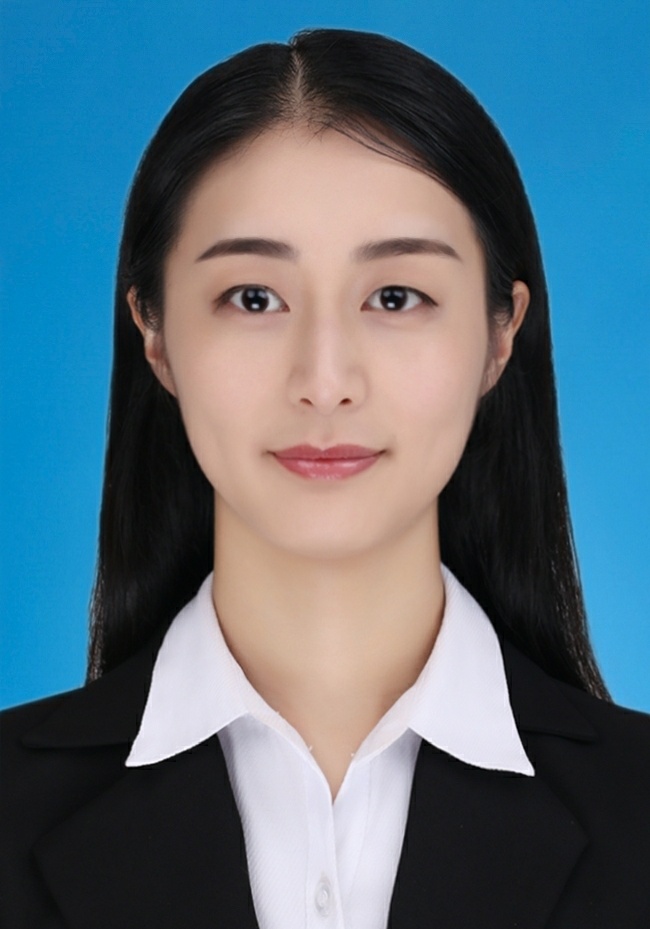}}]{Shiyun Zhao} received the B.E.degree in communication engineering from Yunnan University, Kunming, China, in 2017. She is currently working toward the M.S. degree in Electronics and Communication Engineering at the School of Information Science and Engineering, Shandong University, Qingdao, China. Her current research interests include video transmission and multimedia communication.
\end{IEEEbiography}

\begin{IEEEbiography}
[{\includegraphics[width=1in,height=1.2in,clip,keepaspectratio]{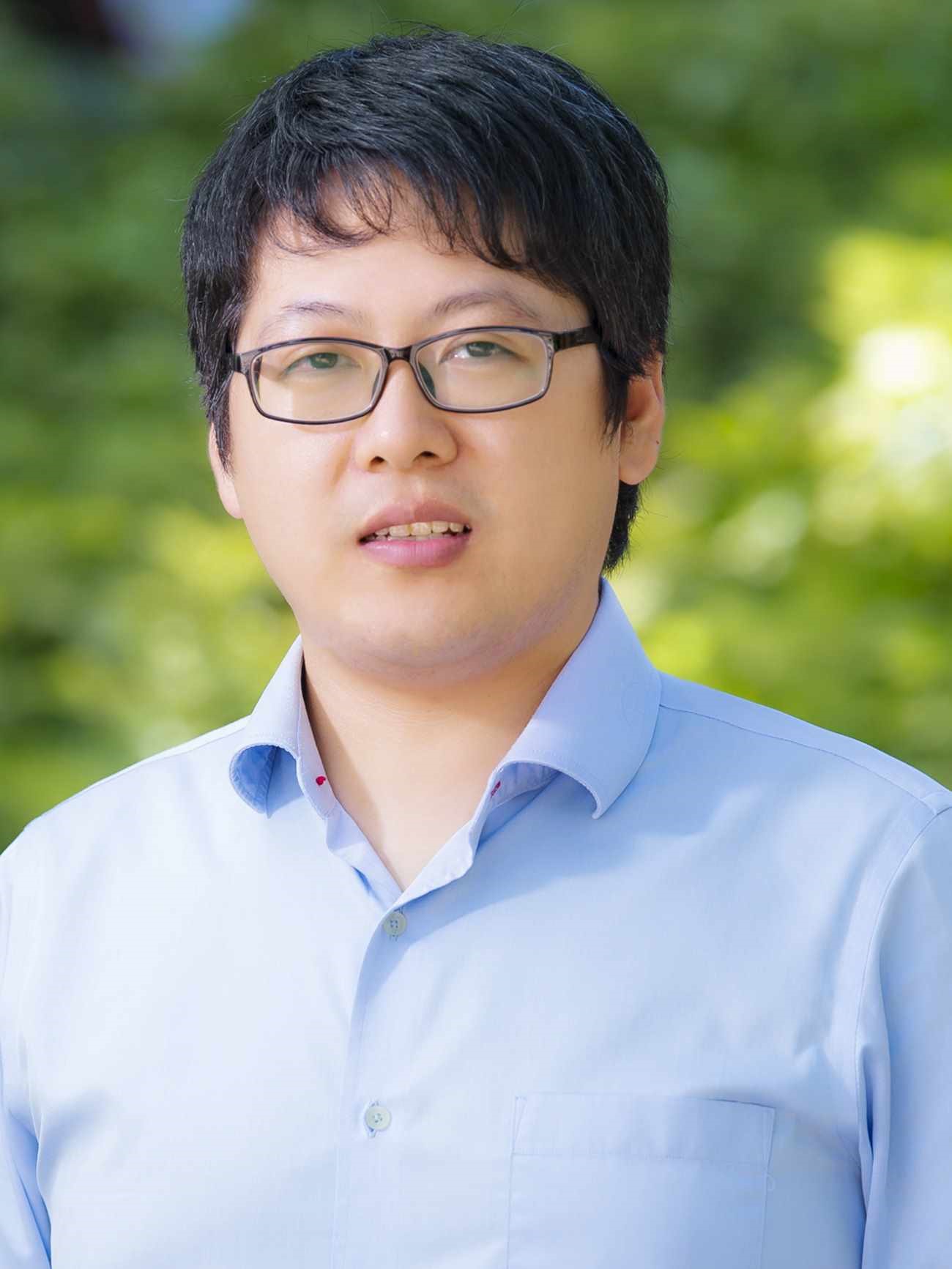}}]{Junhui Hou} (S'13-M'16) received the B.Eng. degree in information engineering (Talented Students Program) from the South China University of Technology, Guangzhou, China, in 2009, the M.Eng. degree in signal and information processing from Northwestern Polytechnical University, Xian, China, in 2012, and the Ph.D. degree in electrical and electronic engineering from the School of Electrical and Electronic Engineering, Nanyang Technological University, Singapore, in 2016. He has been an Assistant Professor with the Department of Computer Science, City University of Hong Kong, since 2017. His research interests include image/video/3D geometry data representation, processing and analysis, semisupervised/unsupervised data modeling for clustering/classification, and data compression and adaptive streaming. Dr. Hou was the recipient of several prestigious awards, including the Chinese Government Award for Outstanding Students Study Abroad from China Scholarship Council in 2015, and the Early Career Award from the Hong Kong Research Grants Council in 2018. He is serving/served as an Associate Editor for The Visual Computer, an Area Editor for Signal Processing: Image Communication, the Guest Editor for the IEEE Journal of Selected Topics in Applied Earth Observations and Remote Sensing and the Journal of Visual Communication and Image Representation, and an Area Chair of ACM International Conference on Multimedia (ACM MM) 2019.
\end{IEEEbiography}

\begin{IEEEbiography}
[{\includegraphics[width=1in,height=1.2in,clip,keepaspectratio]{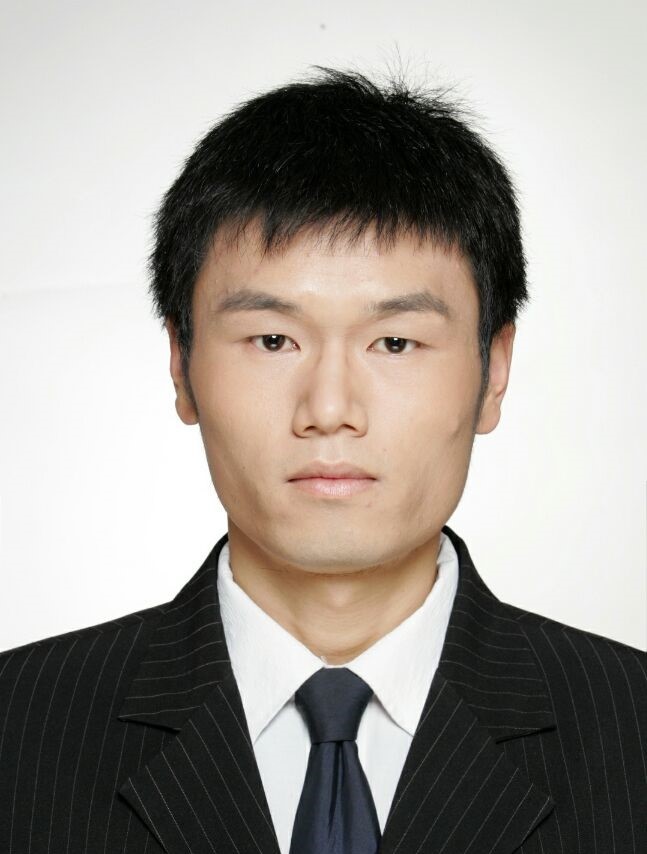}}]{Xuekai Wei} received the B.Eng. degree in Electronic Information Science and Technology from Shandong University in 2014, and the Master's degree in Communication and Information Systems from Shandong University in 2017. He is currently working toward the Ph.D. degree in computer science with the City University of Hong Kong, Hong Kong, China. His current research interests include video coding, video transmission, and machine learning.
\end{IEEEbiography}

\begin{IEEEbiography}
[{\includegraphics[width=1in,height=1.2in,clip,keepaspectratio]{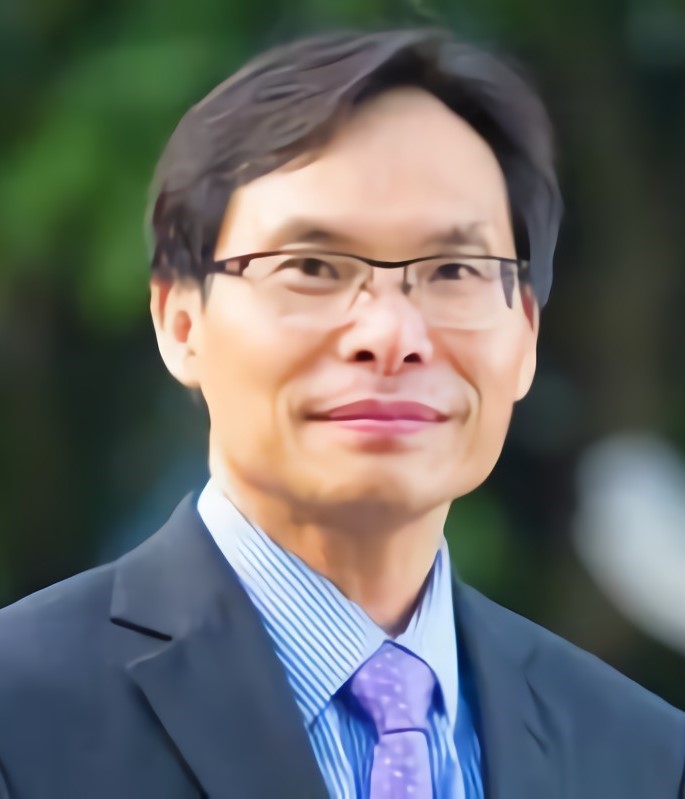}}]{Sam Kwong} (M'93-SM'04-F'13) received the BS degree from the State University of New York at Buffalo, in 1983, the MS degree in electrical engineering from the University of Waterloo, Waterloo, ON, Canada, in 1985, and the PhD degree from the University of Hagen, Germany, in 1996. From 1985 to 1987, he was a diagnostic engineer with Control Data Canada, Mississauga, ON, Canada. He joined Bell Northern Research Canada, Ottawa, ON, Canada, as a member of Scientific Staff. In 1990, he became a lecturer with the Department of Electronic Engineering, The City University of Hong Kong, where he is currently a professor with the Department of Computer Science. His research interests are video and image coding, and evolutionary algorithms. He serves as an associate editor of IEEE Transactions on Industrial Electronics and the IEEE Transactions on Industrial Informatics. He is a fellow of the IEEE.
\end{IEEEbiography}


%\newpage



%\vfill


\end{document}